\documentclass{article} 
\usepackage{iclr2027_conference,times}

\usepackage{amsmath,amsfonts,bm}

\def\eqref#1{equation~\ref{#1}}

\def\1{\bm{1}}

\DeclareMathAlphabet{\mathsfit}{\encodingdefault}{\sfdefault}{m}{sl}
\SetMathAlphabet{\mathsfit}{bold}{\encodingdefault}{\sfdefault}{bx}{n}

\iclrfinalcopy
\usepackage{graphicx}
\usepackage{booktabs}
\usepackage{multirow}
\usepackage{makecell}
\usepackage[table]{xcolor}
\usepackage{algorithm}
\usepackage{algpseudocode}
\usepackage{xspace}

\usepackage{hyperref}

\definecolor{mydarkblue}{rgb}{0,0.08,0.45} 
\definecolor{brickred}{HTML}{9C3B2E}      

\hypersetup{
    colorlinks=true,
    pdfborder={0 0 0},
    citecolor=mydarkblue,
    linkcolor=brickred,
    urlcolor=mydarkblue,
    filecolor=mydarkblue
}

\usepackage{url}
\usepackage{fontawesome5}

\ifdefined\pdfminorversion\pdfminorversion=7\fi

\definecolor{futurecolor}{RGB}{218,233,248}
\newcommand{\ffcell}[1]{\cellcolor{futurecolor}#1}
\newcommand{\method}{\text{FrameFold}\xspace}

\title{\method: Response-Guided KV Cache Consolidation for
Autoregressive Long Video Generation}

\author{
  Muyu Liu$^{1}$\thanks{Work done during an internship at vivo.} \quad
  Mingming Liu$^{2}$ \quad
  Hong Gu$^{2}$ \quad
  Yuyao Zhang$^{1}$ \quad
  Qin Lu$^{2}$\thanks{Corresponding author.} \\
  $^{1}$ShanghaiTech University \\
  $^{2}$vivo BlueImage Lab, vivo Mobile Communication Co., Ltd., China \\
  \href{https://vivocameraresearch.github.io/FrameFold-Webpage/dist}
       {\textbf{Project page}}
}

\begin{document}

\maketitle
\lhead{Preprint}

\begin{abstract}

In autoregressive video generation, retaining all historical key–value (KV) states increases memory usage and per-step attention costs as videos grow longer. A fixed cache budget bounds these costs but raises the challenge of preserving historical content that supports consistent, high-quality generation. Existing methods control cache size through history selection or consolidation, yet deciding what to retain and how to limit the loss of useful content remains a shared challenge. Our analysis shows that spatial queries depend on different historical content and that higher current attention coverage does not necessarily yield better long-video quality. We introduce FrameFold, a training-free KV cache consolidation method that compares the content retrieved by the same recent historical queries before and after merging to guide both representative construction and merge selection. Exploiting temporal redundancy between adjacent historical entries, FrameFold merges their KV states into shared representatives while retaining complete spatial grids. It fits value-merging coefficients in closed form by matching conditional attention responses, then selects the merge with the lowest remaining response distortion. Under matched KV cache budgets, experiments on Self Forcing and Causal Forcing at 30 and 60 seconds and on LongLive at 120 seconds demonstrate improved consistency and visual quality. Continuous eight-minute Self Forcing rollouts further show less quality degradation than the baselines.

\end{abstract}

\section{Introduction}
\label{sec:introduction}

Autoregressive video generation models now synthesize high-quality short
clips \citep{huang2025selfforcing,yin2025causvid,zhu2026causalforcing}.
Although causal generation supports continued prediction of subsequent
frames or chunks \citep{chen2024diffusionforcing,chen2025skyreelsv2,sandai2025magi1,yang2025longlive}, extrapolation beyond the training
horizon often leads to subject and background inconsistencies and a decline
in visual quality
\citep{li2026rollingsink,li2025stablevideoinfinity,liu2025rollingforcing,yi2025deepforcing}.
Historical keys and values (KV) provide context for consistent subsequent
generation \citep{chen2026contextforcing,hu2026longliverag,zhu2025memorizeandgenerate}, but retaining the full
history makes storage and per-step attention costs grow with video length
\citep{lu2026fademem}. Under a fixed cache budget, the model must repeatedly
discard or compress history, determining what information remains available
to later generation \citep{kim2026memrope,lu2026fademem,yi2025deepforcing,zhao2026densitykv}.
We study how to organize and compress useful history within this budget to sustain long-video consistency and visual quality.

Existing methods manage limited KV caches through token selection and memory
consolidation. Deep Forcing retains historical tokens based on attention salience
\citep{yi2025deepforcing}, potentially discarding cues needed for subsequent
generation. FlowCache improves coverage through redundancy-aware scoring
\citep{ma2026flowcache}, while Future Forcing uses future-query proxies and
merges values from some evicted tokens into retained tokens to reduce
information loss caused by eviction  \citep{luo2026futureforcing}. For memory consolidation,
MemRoPE updates memories through exponential moving averages
\citep{kim2026memrope}, while FadeMem merges adjacent historical entries
according to temporal distance \citep{lu2026fademem}, extending temporal
coverage within a fixed memory budget. However, both methods use preset update
rates or temporal structure to determine update weights or merge choices,
without directly evaluating the resulting content loss. Token selection and
memory consolidation thus share a challenge: assessing the value of history for subsequent generation to decide \textbf{what to retain and how to retain it.}

\begin{figure}[ht]
    \centering
    \includegraphics[width=\linewidth]{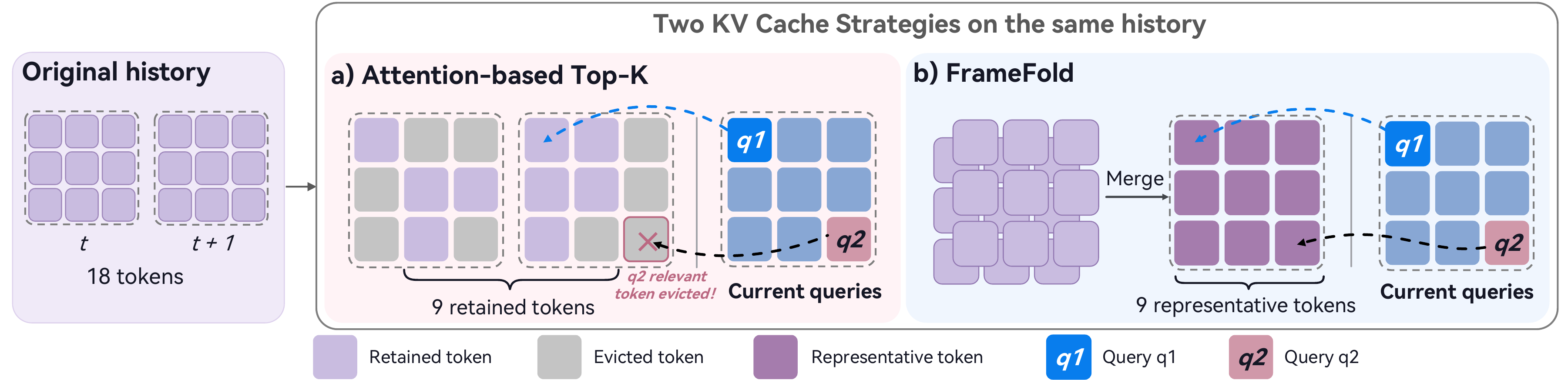}
    \caption{\textbf{Token selection and shared representation.}
    (a) Attention-based Top-K selection retains subsets of historical tokens;
    (b) \method consolidates adjacent history into a representative grid.
    Actual Deep Forcing selection maps are shown in
    Appendix~\ref{app:selection-maps}.}
    \label{fig:teaser}
\end{figure}

Our analysis shows that spatial queries depend on different historical
content, and a shared Top-K ranking based on aggregated attention can omit
information important to some queries (Figure~\ref{fig:motivation}(a)). At a
matched candidate retention ratio, replacing Deep Forcing's Top-K rule with
Bottom-K reduces current attention coverage yet improves the evaluated
consistency and visual-quality metrics on two generators, with
lower dynamic degree (Figure~\ref{fig:motivation}(b)). This reversal shows
that current attention coverage alone is insufficient to assess history
retention. To meet different queries' historical needs within a fixed
budget, we exploit temporal redundancy in neighboring frames and their
features \citep{ji2026forcingkv,lu2026fademem,samuel2026fastar,xi2026quantvideogen}
to merge adjacent historical KV entries at corresponding spatial locations.
Each representative retains a complete spatial grid while covering a longer
historical interval. The key is to limit content loss so that shared
representatives preserve as closely as possible what the original history
supplies to different queries.

We introduce \method, a training-free KV cache consolidation method that uses
conditional-response preservation to guide representative construction and
merge selection (Figure~\ref{fig:teaser}). Recent historical queries serve as
probes to compare the content retrieved from the original history and its
shared representative. Normalizing attention within each block separates
this comparison from the history's current share of global attention. For
each candidate pair, \method merges keys by temporal span and fits the
value-merging coefficient in closed form to match the original conditional
response. It then executes the merge with the lowest remaining response
distortion.


Under matched KV cache budgets, FrameFold achieves the highest consistency, motion smoothness, and visual-quality scores among the evaluated methods on Self Forcing and Causal Forcing at both 30 and 60 seconds. Experiments on LongLive at 120 seconds further demonstrate gains in consistency and visual quality. 
Continuous eight-minute Self Forcing rollouts show higher consistency and visual quality with smaller declines than those of FadeMem and Deep Forcing.
Ablation studies support conditional-response matching as a consolidation objective and the contributions of representative construction and merge selection.

\section{Related Work}
\label{sec:related-work}

\textbf{Autoregressive Video Generation.}
Autoregressive video generation predicts frames or chunks sequentially,
providing a basis for streaming synthesis
\citep{chen2024diffusionforcing,chen2025skyreelsv2,sandai2025magi1}.
Recent work improves generation quality through advances in training and
distillation
\citep{guo2025resamplingforcing,huang2025selfforcing,yin2025causvid,zhu2026causalforcing}
and enhances long-horizon stability through training and rollout strategies
\citep{cui2025selfforcingpp,li2025stablevideoinfinity,liu2025rollingforcing,yang2025longlive}.
Autoregressive generation also supports interactive world modeling under
action, camera, or text control
\citep{lingbot-world-v2,huang2026vid2world,mao2025yume15}.
To support consistent scene revisits, WorldPlay
\citep{sun2026worldplay} retrieves and reorganizes historical context,
while Matrix-Game~3.5 \citep{qian2026matrixgame35} uses geometry-aware
patch memory.
As generated sequences grow longer, retaining and organizing useful history
within a limited KV cache becomes important for maintaining long-term
consistency.

\textbf{KV cache selection.}
KV cache selection determines which historical states to retain and how to
allocate limited capacity. In language models, retention policies exploit
historical attention contributions and the persistence of importance
\citep{liu2023scissorhands,zhang2023h2o}, or preserve attention sinks and
recent context \citep{xiao2024streamingllm}. Further work adapts retention
to head-specific attention patterns \citep{ge2024fastgen,li2024snapkv}
and allocates cache capacity across layers \citep{cai2025pyramidkv}.
In autoregressive video generation, selection criteria include
attention-based importance \citep{yi2025deepforcing}, learned salience
\citep{chen2026pafukv}, predicted future-query relevance
\citep{luo2026futureforcing}, and temporal redundancy
\citep{ma2026flowcache}. Relax Forcing \citep{zhao2026relaxforcing}
balances global stability against redundancy relative to recent context,
while TetherCache \citep{meng2026tethercache} combines attention relevance
with temporal coverage of historical frames and aligns the statistics of
recalled memory. Beyond selection criteria, retention and cache allocation adapt to the context needs of individual frames and attention heads~\citep{cai2026focusedforcing,guo2026dummyforcing,ji2026forcingkv}.

\textbf{History Compression and Response Preservation.}
Complementary to selection, history compression encodes past information
into compact representations. FramePack packs historical frames into
compact context \citep{zhang2025framepack}, TinyHistory learns lightweight
history embeddings \citep{zhang2025tinyhistory}, and MAG
\citep{zhu2025memorizeandgenerate} and Echo-Infinity
\citep{bian2026echoinfinity} aggregate history through learned memories.
Representation merging offers another compression route: ToMe and VidToMe
apply similarity-based token merging to vision transformers and video
editing, respectively \citep{bolya2023tome,li2023vidtome}.
In language models, MiniCache exploits KV similarity between adjacent
layers for compression across network depth \citep{liu2024minicache},
while KVMerger merges KV states according to key similarity
\citep{wang2024kvmerger}. In autoregressive video generation, MemRoPE
aggregates history through exponential moving averages
\citep{kim2026memrope}, and FadeMem merges adjacent complete spatial grids
\citep{lu2026fademem}. TempCache uses cross-frame correspondences for
compression \citep{samuel2026fastar}, while Future Forcing guides merging
with future-query projection profiles \citep{luo2026futureforcing}.
Response preservation also guides representation construction: Attention
Matching compacts language-model KV caches by matching local attention
outputs and unnormalized attention mass
\citep{zweiger2026attentionmatching}. Like FadeMem, \method merges adjacent
complete spatial grids. It constructs shared representatives through
conditional-response matching and selects merges by the remaining response
distortion after fitting.

\section{Motivation}
\label{sec:problem-formulation}

\begin{figure}[t]
    \centering
    \includegraphics[width=\linewidth]{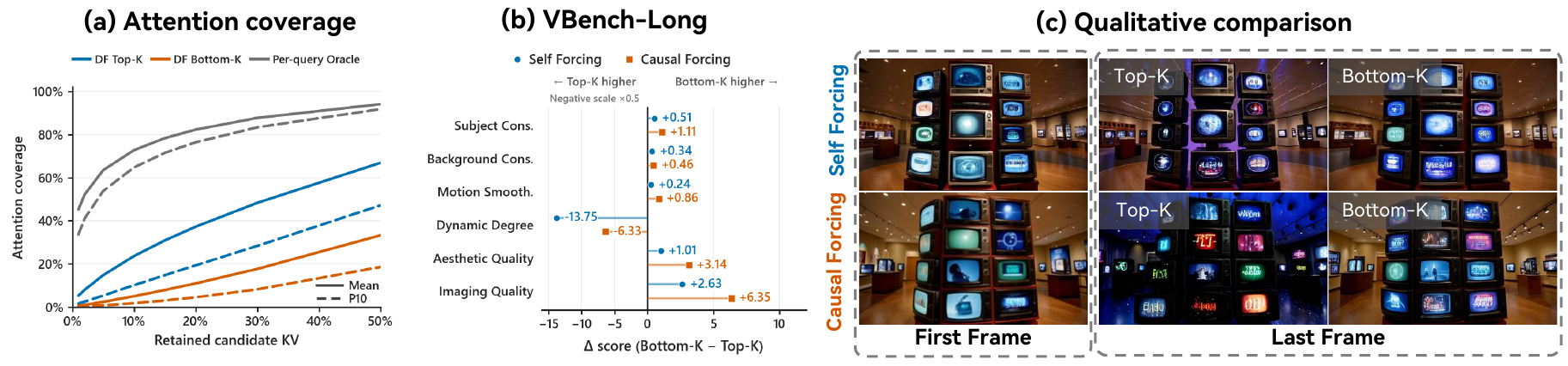}
    \caption{\textbf{Shared Top-K leaves query-level coverage gaps; higher
    coverage need not yield better generation quality.}
    \textbf{(a)} Candidate-region attention coverage for Top-K, Bottom-K,
    and a per-query oracle (solid: mean; dashed: P10).
    \textbf{(b)} VBench-Long score differences (Bottom-K minus Top-K).
    Despite lower coverage, Bottom-K improves five consistency and quality
    metrics on both models, with lower dynamic degree.
    \textbf{(c)} First-frame references and final frames for Self Forcing
    (top) and Causal Forcing (bottom).
    See Appendix~\ref{app:topk-bottomk-visual} for experimental details.}
    \label{fig:motivation}
\end{figure}

\subsection{Attention Coverage and Generation Quality}
\label{sec:attention-retention-limits}

Queries at different spatial locations in a video often attend to different
historical content. Even when an individual query concentrates its
attention on a small subset of historical tokens, the subsets required by
different queries need not coincide
\citep{cai2026focusedforcing,guo2025resamplingforcing}. Under a fixed cache
budget, Deep Forcing aggregates attention from multiple recent queries into
an importance score for each historical token and retains the Top-K tokens
according to this shared ranking \citep{yi2025deepforcing}
(Figure~\ref{fig:teaser}(a)). This policy serves different queries with
the same retained subset: history that is highly relevant to a few queries
may rank below history that is broadly relevant to many queries and
consequently be discarded.

We examine this issue through query-level attention coverage analysis
(Figure~\ref{fig:motivation}(a)). At 50\% candidate token retention, shared
Top-K retains roughly two-thirds of the candidate-region attention on
average, while its 10th-percentile coverage (P10) falls below one-half. By
contrast, a per-query oracle that independently selects tokens for each
query exceeds 90\% in both mean coverage and P10 at the same retention
ratio. The larger gap at P10 highlights the limited retention of history
attended to by some queries. Visualizations of actual token selection also
show uneven retention frequencies across spatial locations
(Figure~\ref{fig:topk-selection-maps} in the appendix).

Higher attention coverage does not necessarily lead to better subsequent
generation. We replace Top-K with Bottom-K while keeping the scoring
procedure and candidate retention ratio unchanged. Despite retaining less
current attention, Bottom-K achieves higher scores on all five consistency
and quality metrics on both Self Forcing and Causal Forcing, with lower
dynamic degree (Figure~\ref{fig:motivation}(b)). In the Causal Forcing
example, Bottom-K also better preserves the initial scene's colors, whereas
Top-K exhibits a pronounced blue cast (Figure~\ref{fig:motivation}(c)). These
results show that lower-ranked history remains useful for generation.
Compression decisions under a fixed budget should therefore consider \textbf{what
historical content different queries can still retrieve after compression;
current attention share alone is insufficient to assess the value of
retaining that content.}

\subsection{From Token Selection to Shared Representatives}
\label{sec:specified-reconstructibility}

To preserve history that different queries may need within a fixed budget,
we consider allowing multiple historical entries to share cache slots.
Temporal redundancy in video offers this opportunity: QVG
\citep{xi2026quantvideogen} reports that latent tokens at corresponding
spatial locations in adjacent frames often exhibit high similarity,
suggesting that overlapping historical content can be stored through
shared representatives. To accommodate the retrieval needs of queries at
different spatial locations, we retain complete spatial grids and merge
adjacent historical entries along the temporal dimension, combining KV features at
corresponding spatial locations into a shared representative
(Figure~\ref{fig:teaser}(b)). Each spatial location thus retains a
historical representation, while a single grid represents a longer
contiguous interval of history, reducing the number of stored entries. The
remaining challenge is to control the content loss from merging so that
the shared representative faithfully preserves what the original history
supplies to queries.

\section{Method}
\label{sec:method}

\subsection{Evaluating Merges through Conditional Responses}
\label{sec:conditional-consolidation}

History supplies content to queries through attention readouts: a query
uses keys to form retrieval weights and aggregates values accordingly.
We therefore assess content preservation by comparing the representations
read by the same query before and after merging. In full-cache attention,
a historical block's contribution to the output is also scaled by its
share of global attention. To separate the block's readout from this share,
we normalize attention within the block being evaluated and refer to the
resulting weighted sum of values as its \emph{conditional response}.

We measure the readout distortion introduced by merging as the difference
between the conditional responses of the original history and its shared
representative under the same query. Content preservation under this
criterion leads to two coupled decisions: \textbf{which pair to merge}
to minimize the resulting response distortion
(Section~\ref{sec:efficient-scoring}); and
\textbf{how to construct the shared representative} to match the
original history's conditional responses as closely as possible
(Section~\ref{sec:structured-memory}).

\begin{figure}[ht]
    \centering
    \includegraphics[width=\linewidth]{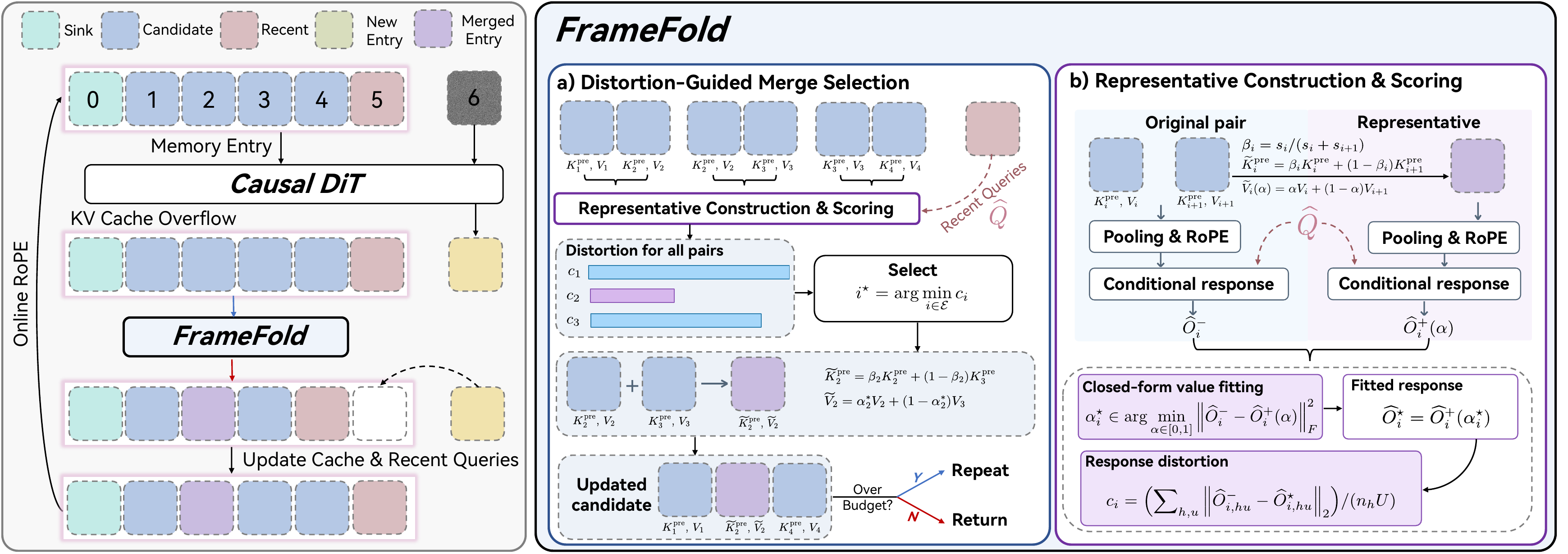}
    \caption{\textbf{Overview of \method.}
    When the KV cache exceeds its budget, \method consolidates adjacent
    entries in the candidate history while protecting sink and recent
    entries. Recent historical queries guide the evaluation of candidate
    merges, and the selected pair is replaced by a shared KV representative.
    Here, $s_i$ denotes the number of original latent frames represented
    by entry $i$.}
    \label{fig:pipeline}
\end{figure}

\subsection{Distortion-Guided Merge Selection}
\label{sec:efficient-scoring}

When the KV cache exceeds its budget, we protect the initial sink entries
and recent history and select adjacent entries in the intermediate
candidate region for merging. Suppose this region currently contains $n$
historical entries $\mathcal F_1,\ldots,\mathcal F_n$ in temporal order.
There are $n-1$ adjacent candidate pairs, indexed by
$\mathcal E=\{1,\ldots,n-1\}$, where candidate $i$ corresponds to
$(\mathcal F_i,\mathcal F_{i+1})$.
Each iteration compares these candidates and executes one merge.

We use merge matrices to express the transformation from each candidate
pair to its shared representative. Suppose each entry contains $S$ spatial
tokens. We denote the keys of entry $\mathcal F_i$ before rotary position
encoding (RoPE) by $K_i^{\mathrm{pre}}$ and the corresponding values by
$V_i$. Merging operates on the pre-RoPE keys and the corresponding values:
\begin{equation}
    \widetilde K_i^{\mathrm{pre}}
    =M_i^K[K_i^{\mathrm{pre}};K_{i+1}^{\mathrm{pre}}],
    \qquad
    \widetilde V_i=M_i^V[V_i;V_{i+1}],
    \label{eq:candidate-merge-matrices}
\end{equation}
where $M_i^K,M_i^V\in\mathbb R^{S\times2S}$ map the two key and value
grids, respectively, into one shared grid, and $[\,;\,]$ denotes
concatenation along the token dimension. For each candidate, we first
determine the merge matrices and construct a temporary representative using the rule in
Section~\ref{sec:structured-memory}, and then evaluate its conditional-response
distortion relative to the original pair.

For response evaluation, we denote queries after RoPE by $Q$. We obtain
$K_i$ by applying RoPE to $K_i^{\mathrm{pre}}$ at the entry's current cache
positions, and $\widetilde K_i$ by applying RoPE to
$\widetilde K_i^{\mathrm{pre}}$ at the representative's positions.
Values do not receive RoPE. Hatted quantities denote evaluation
representations derived from these QKV features; their construction is
described in Section~\ref{sec:structured-memory}. All candidates share the
same query probes $\widehat Q$. The original pair and its fitted
representative have conditional responses
\begin{equation}
    \begin{aligned}
    \widehat O_i^-
    &=\operatorname{softmax}\!\bigl(
        \widehat Q[\widehat K_i;\widehat K_{i+1}]^\top/\sqrt d
      \bigr)[\widehat V_i;\widehat V_{i+1}],\\
    \widehat O_i^\star
    &=\operatorname{softmax}\!\bigl(
        \widehat Q\widehat{\widetilde K}_i^\top/\sqrt d
      \bigr)\widehat{\widetilde V}_i,
    \end{aligned}
    \label{eq:pooled-conditional-responses}
\end{equation}
where $d$ is the attention head dimension. These responses describe what
the same queries read from the original pair and the shared representative,
with each softmax normalized within
its respective historical block. We define the merge cost as their mean
response difference:
\begin{equation}
    c_i=\Bigl(
        \sum_{h,u}
        \left\|
            \widehat O^-_{i,hu}
            -\widehat O^\star_{i,hu}
        \right\|_2
    \Bigr)/(n_h U),
    \label{eq:proxy-merge-score}
\end{equation}
where the sum runs over $n_h$ attention heads and $U$ query probes, and
each $L_2$ norm is taken over value channels. Every merge replaces two
complete spatial grids with one, releasing the same cache capacity. We
therefore select the candidate with the lowest cost:
\begin{equation}
    i^\star=\arg\min_{i\in\mathcal E}c_i.
    \label{eq:merge-selection}
\end{equation}
Each layer independently scores candidates and selects a pair shared by
all its heads. We apply the selected merge matrices to the complete
$K^{\mathrm{pre}}$ and $V$ grids and replace the pair with a representative
carrying their combined temporal span. We then reconstruct representatives
and rescore all current
candidates using the updated history, repeating until the cache meets its
budget; query probes remain fixed throughout the update.

\subsection{Response-Guided Representative Construction}
\label{sec:structured-memory}

We construct a shared representative for each candidate pair
scored in Section~\ref{sec:efficient-scoring}. Given a fixed adjacent pair
$(\mathcal F_i,\mathcal F_{i+1})$ and query probes, we determine its merge
matrices so that the representative approximates the original conditional
responses.

To preserve spatial organization, we form convex combinations of the two
entries' features at corresponding locations. To keep fitting inexpensive,
keys and values each use one coefficient shared across spatial locations
and attention heads:
\begin{equation}
    M_i^K=\bigl[\beta_i I_S,\;(1-\beta_i)I_S\bigr],
    \qquad
    M_i^V(\alpha)=\bigl[\alpha I_S,\;(1-\alpha)I_S\bigr],
    \label{eq:full-resolution-merge}
\end{equation}
where $I_S$ is the identity matrix and $\beta_i,\alpha\in[0,1]$.
We determine these two coefficients in turn.

Keys affect conditional responses nonlinearly through softmax. We first
construct the shared keys $\widetilde K_i^{\mathrm{pre}}$ before RoPE
\citep{su2021roformer} using the span-weighted averaging rule of
FadeMem (v1)~\citep{lu2026fademem}. If entry
$\mathcal F_i$ covers $s_i$ original latent frames, we set
$\beta_i=s_i/(s_i+s_{i+1})$, weighting the keys by the number of frames
each entry represents.

With the representative's keys and positions fixed, its attention weights
$A_i^+=\operatorname{softmax}\!\bigl(
\widehat Q\widehat{\widetilde K}_i^\top/\sqrt d\bigr)$
are independent of $\alpha$. Its response is affine in $\alpha$:
\mbox{$\widehat O_i^+(\alpha)=C_i+\alpha D_i$}, where
$C_i=A_i^+\widehat V_{i+1}$ and
$D_i=A_i^+(\widehat V_i-\widehat V_{i+1})$.
We fit this candidate's coefficient by minimizing the squared response
error against the original response $\widehat O_i^-$:
\begin{equation}
    \alpha_i^\star
    \in\arg\min_{\alpha\in[0,1]}
    \left\|
        \widehat O_i^- - C_i - \alpha D_i
    \right\|_F^2.
    \label{eq:value-fit-objective}
\end{equation}
This objective is a scalar convex quadratic. For $\|D_i\|_F^2>0$, its
interval-constrained optimum is
\begin{equation}
    \alpha_i^\star
    =\operatorname{clip}_{[0,1]}\!\bigl(
        \langle D_i,\widehat O_i^- - C_i\rangle_F
        /\|D_i\|_F^2
      \bigr).
    \label{eq:value-fit-solution}
\end{equation}
Under the same shared keys and query probes, the span-weighted coefficient
$\alpha=\beta_i$ is feasible, so the fitted values incur no greater squared
response error than span-weighted values. Substituting $\alpha_i^\star$
into $M_i^V$ gives the candidate representative and its response
$\widehat O_i^\star=\widehat O_i^+(\alpha_i^\star)$ for computing the merge
cost in Section~\ref{sec:efficient-scoring}.

To reduce the cost of representative fitting and candidate scoring, we
construct the hatted evaluation representations through pooling. We
jointly pool saved queries from the latest $R$ completed historical frames
over time and space into $U$ probes, and spatially pool each historical
entry's keys and values to $U$ tokens.
Pooling operates on the position-encoded queries $Q$ and keys $K_i$,
and directly on values $V_i$.
Fitting and scoring use these pooled representations, and the selected
mixing coefficients are applied to the complete $K^{\mathrm{pre}}$ and $V$
spatial grids. Additional position conventions and degenerate-case handling are given in
Appendix~\ref{app:efficient-implementation}.

\section{Experiments}
\label{sec:experiments}

\subsection{Experimental setup}
\label{sec:experimental-setup}

We evaluate \method on pretrained Self Forcing and Causal Forcing models
\citep{huang2025selfforcing,zhu2026causalforcing} without additional training.
The main comparison uses 128 MovieGenBench prompts
\citep{polyak2024moviegen} refined with Qwen2.5-7B-Instruct, following
the prompt setup of MemRoPE~\citep{kim2026memrope}. We generate 30- and
60-second videos at $480\times832$ resolution and 16 fps. Baselines comprise the original
backbones with FIFO cache eviction, Deep Forcing, and FadeMem
\citep{lu2026fademem,yi2025deepforcing}. Within each generator--duration
setting, methods share checkpoints, prompts, random seeds, and denoising
schedules. All methods use a default budget of 12 frame-equivalent KV entries,
including the three-frame current chunk, and protect three sink frames and
the three most recent historical frames. Unless varied explicitly, \method
uses a spatial pooling factor of 4 and queries from the latest $R=3$
completed historical frames for representative fitting and merge scoring.


We report the six VBench-Long dimensions \citep{huang2024vbench,huang2026vbenchpp} supported for custom video evaluation: subject consistency and background consistency for temporal consistency; motion smoothness and dynamic degree for motion behavior; and aesthetic quality and imaging quality for visual quality. Scores are reported on a 0--100 scale.

\subsection{Main results}
\label{sec:main-results}

\begin{table}[t]
\caption{
\textbf{Quantitative comparison of \method with baselines on 30- and
60-second video generation.}
We evaluate Self Forcing and Causal Forcing on 128 MovieGenBench prompts
with a cache budget of 12 frame-equivalent KV entries.
Higher is better; best scores within each backbone--duration group are
in \textbf{bold}.
}
\label{tab:main-results}

\setlength{\tabcolsep}{1.9pt}

\begin{center}
\begin{tabular}{
    c
    l
    c c c c c c
}
\toprule

\textbf{Config}
&
\textbf{Method}
&
\makecell{\textbf{Subject}\\\textbf{Consist.}$\uparrow$}
&
\makecell{\textbf{Background}\\\textbf{Consist.}$\uparrow$}
&
\makecell{\textbf{Motion}\\\textbf{Smooth.}$\uparrow$}
&
\makecell{\textbf{Dynamic}\\\textbf{Degree}$\uparrow$}
&
\makecell{\textbf{Aesthetic}\\\textbf{Quality}$\uparrow$}
&
\makecell{\textbf{Imaging}\\\textbf{Quality}$\uparrow$}
\\

\midrule


\multirow[c]{8}{*}{30s}
&
Self Forcing
                    & 96.38 & 95.89 & 97.77 & 59.11  
                    & 57.40 & 66.80\\
&
+ Deep Forcing
                    & 96.72 & 96.03 & 98.09 & 56.67 
                    & 57.84 & 67.27\\
&
+ FadeMem
                    & 96.94 & 96.15 & 98.08 & 56.35  
                    & 58.01 & 68.29\\
&
\ffcell{\textbf{+ FrameFold}}
                    & \ffcell{\textbf{97.16}} & \ffcell{\textbf{96.29}} & \ffcell{\textbf{98.25}} &  \ffcell{\textbf{60.42}} 
                    & \ffcell{\textbf{58.27}} & \ffcell{\textbf{68.52}}\\

\cmidrule(lr){2-8}

&
Causal Forcing
                    & 95.65 & 95.52 & 96.95 & 75.73  
                    & 55.80 & 66.31\\

&
+ Deep Forcing
                    & 95.53 & 95.48 & 96.93 & 78.59  
                    & 56.35 & 65.53\\
&
+ FadeMem
                    & 95.73 & 95.44 & 97.08 & \textbf{83.85}  
                    & 57.08 & 68.08\\
&
\ffcell{\textbf{+ FrameFold}}
& \ffcell{\textbf{95.82}}
& \ffcell{\textbf{95.53}}
& \ffcell{\textbf{97.17}}
& \ffcell{83.13}
& \ffcell{\textbf{57.55}}
& \ffcell{\textbf{68.91}}
\\

\midrule


\multirow[c]{8}{*}{60s}
&
Self Forcing
                    & 95.80 & 95.60 & 97.49 & 55.34  
                    & 56.95 & 66.12\\

&
+ Deep Forcing
                    & 96.39 & 95.89 & 98.02 & 57.59 
                    & 57.19 & 67.14\\

&
+ FadeMem
                    & 96.71 & 96.05 & 98.14 & 54.17  
                    & 57.80 & 68.17\\

&
\ffcell{\textbf{+ FrameFold}}
                    & \ffcell{\textbf{96.94}} & \ffcell{\textbf{96.17}} & \ffcell{\textbf{98.23}} & \ffcell{\textbf{62.11}}  
                    & \ffcell{\textbf{58.23}} & \ffcell{\textbf{68.80}}\\

\cmidrule(lr){2-8}

&
Causal Forcing
                    & 94.68 & 95.12 & 96.12 & 75.07  
                    & 54.85 & 64.54\\

&
+ Deep Forcing
                    & 95.45 & 95.16 & 96.85 & 73.67  
                    & 55.64 & 64.87\\
&
+ FadeMem
                    & 95.38 & 95.23 & 96.76 & \textbf{85.31}  
                    & 56.03 & 68.09\\
&
\ffcell{\textbf{+ FrameFold}}
& \ffcell{\textbf{95.54}}
& \ffcell{\textbf{95.36}}
& \ffcell{\textbf{97.09}}
& \ffcell{83.36}
& \ffcell{\textbf{56.70}}
& \ffcell{\textbf{68.35}}
\\

\bottomrule
\end{tabular}
\end{center}
\end{table}

\begin{figure}[!t]
    \centering
    \includegraphics[width=\linewidth]{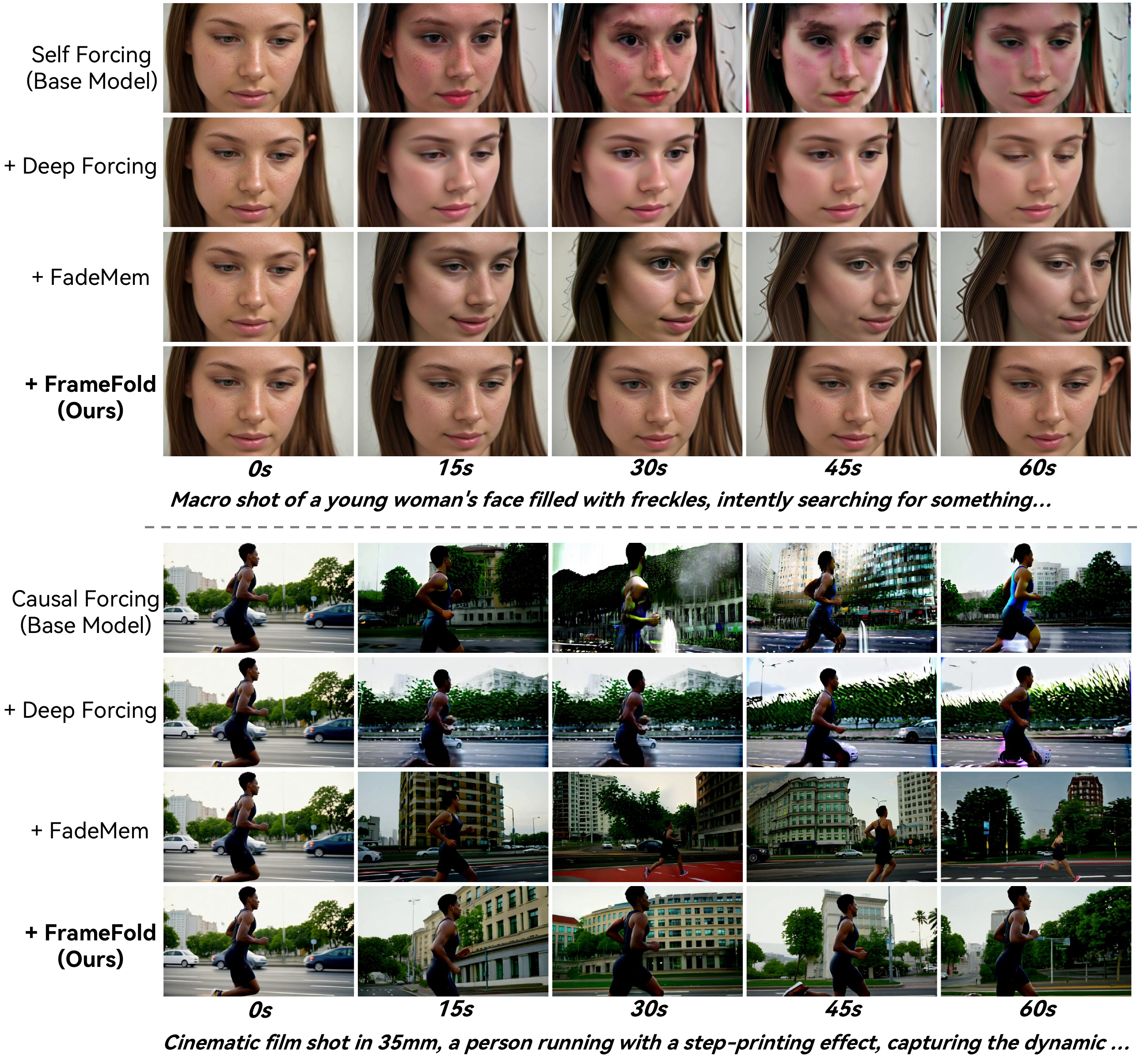}
    \caption{\textbf{Qualitative comparison of \method with baselines on
    60-second video generation.}
    Self Forcing and Causal Forcing examples are shown in the top and bottom
    groups, respectively. Methods within each example use the same prompt
    and seed, with a cache budget of 12 frame-equivalent KV entries.}
    \label{fig:main-results-visual}
\end{figure}

\method improves consistency and visual quality across both generators and
rollout lengths under the same cache budget (Table~\ref{tab:main-results}).
It achieves the best or tied-best subject consistency, background
consistency, motion smoothness, aesthetic quality, and imaging quality in
all four settings. At 60 seconds, aesthetic quality exceeds that of the strongest
baseline by 0.43 points on Self Forcing and 0.67 points on Causal Forcing.
Dynamic degree is highest on Self Forcing and lower than that of FadeMem on
Causal Forcing, indicating that the quality gains accompany different
motion levels across the two backbones.
Appendix~\ref{app:efficiency} compares generation throughput at the shared
12-entry budget and additionally evaluates FIFO and Deep Forcing with
21 entries.

In the portrait example in Figure~\ref{fig:main-results-visual}, \method retains
visible freckles and more stable skin coloration, while Deep Forcing loses
fine facial texture and FIFO develops a stronger color cast. In the running
example, Deep Forcing exhibits fragmented background textures in later
frames, whereas \method maintains cleaner scene structure and subject
boundaries.

\subsection{Long-horizon generation}
\label{sec:long-horizon}

We test sustained compression by generating continuous 480-second videos
with Self Forcing on 24 prompts, using the same fixed cache budget. For each
method, the evaluated 30-, 60-, 120-, 240-, and 480-second videos are nested
prefixes of the same rollout.

\begin{figure}[t]
    \centering
    \includegraphics[width=\linewidth]{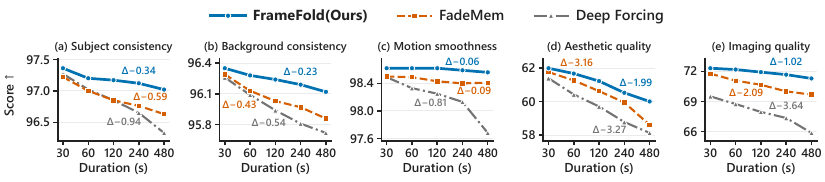}
    \caption{\textbf{Quantitative comparison of \method with baselines on
    video generation up to 480 seconds.}
    We evaluate Self Forcing on 24 prompts with a cache budget of 12
    frame-equivalent KV entries, using nested prefixes of continuous
    rollouts. Colored labels show
    $\Delta=s_{480\mathrm{s}}-s_{30\mathrm{s}}$; the duration axis uses a
    base-2 log scale.}
    \label{fig:long-duration}
\end{figure}

Figure~\ref{fig:long-duration} shows higher subject and background
consistency, aesthetic quality, and imaging quality for \method than for
FadeMem and Deep Forcing at all five durations. The decline from 30 to
480 seconds is also smaller in each of these four dimensions. At 480 seconds,
\method exceeds FadeMem by 1.40 points in aesthetic quality and 1.61 points
in imaging quality.
Complete six-dimensional results (Table~\ref{tab:ab-duration}) and temporal
visualizations appear in Appendix~\ref{app:long-horizon}.

\subsection{Ablation studies}
\label{sec:ablations}

Merge-selection and value-fitting ablations use 48 prompts, and the
response-objective comparison uses 128 prompts. These studies use
Self Forcing with 30-second videos and the default 12-entry cache budget.
\begin{table}[!htbp]
\caption{\textbf{Ablation on merge selection and value fitting.}
Self Forcing on 48 prompts. K/V/O diff measure key/value/conditional-response
differences; Span/Optim use span-weighted/fitted values.
Best scores are in \textbf{bold}.}
\label{tab:ablation-merge-v}
\setlength{\tabcolsep}{1.9pt}
\begin{center}
\begin{tabular}{ll | cccccc}
\toprule
\textbf{Selection} & \makecell{\textbf{Value}\\\textbf{merging}}
& \makecell{\textbf{Subject}\\\textbf{Consist.}$\uparrow$}
& \makecell{\textbf{Background}\\\textbf{Consist.}$\uparrow$}
& \makecell{\textbf{Motion}\\\textbf{Smooth.}$\uparrow$}
& \makecell{\textbf{Dynamic}\\\textbf{Degree}$\uparrow$}
& \makecell{\textbf{Aesthetic}\\\textbf{Quality}$\uparrow$}
& \makecell{\textbf{Imaging}\\\textbf{Quality}$\uparrow$}\\
\midrule
\multirow{2}{*}{V diff}
& Span    & 97.31 & 96.26 & 98.17 & \textbf{62.36} & 60.33 & 70.58\\
& Optim   & 97.37 & 96.31 & 98.19 & 60.41 & 60.39 & 70.57\\
\midrule
\multirow{2}{*}{K diff}
& Span    & 97.40 & 96.25 & 98.17 & 61.11 & 60.34 & 70.56\\
& Optim   & 97.45 & 96.33 & 98.20 & 59.44 & 60.38 & 70.69\\
\midrule
\multirow{2}{*}{\textbf{O diff}}
& Span    & 97.48 & 96.33 & 98.25 & 57.64 & \textbf{60.73} & 70.67\\
& Optim   & \textbf{97.66} & \textbf{96.43} & \textbf{98.28}
          & 59.31 & 60.66 & \textbf{70.80}\\
\bottomrule
\end{tabular}
\end{center}
\end{table}

\textbf{Merge selection.}
With either Span or Optim, O diff achieves the highest subject and
background consistency, motion smoothness, aesthetic quality, and imaging
quality among the three selection criteria, with lower dynamic degree
(Table~\ref{tab:ablation-merge-v}).

\textbf{Value fitting.}
Replacing Span with Optim improves subject and background consistency and
motion smoothness under all three selection criteria. Imaging quality
improves with K diff and O diff, with a 0.01-point decrease under V diff.
With O diff, fitting increases subject consistency by 0.18 points and
imaging quality by 0.13 points, with a 0.07-point decrease in aesthetic
quality. Qualitative comparisons appear in Appendix~\ref{app:merge-visual}.

\begin{table}[!htbp]

\caption{\textbf{Ablation on the response objective.}
Self Forcing on 128 prompts. Each response objective is used for both
value fitting and merge selection.
Best scores are in \textbf{bold}.}
\label{tab:ablation-response}

\setlength{\tabcolsep}{2.5pt}

\begin{center}
\begin{tabular}{
    l
    c c c c c c
}
\toprule

\makecell[l]{\textbf{Response}\\\textbf{objective}}
&
\makecell{\textbf{Subject}\\\textbf{Consist.}$\uparrow$}
&
\makecell{\textbf{Background}\\\textbf{Consist.}$\uparrow$}
&
\makecell{\textbf{Motion}\\\textbf{Smooth.}$\uparrow$}
&
\makecell{\textbf{Dynamic}\\\textbf{Degree}$\uparrow$}
&
\makecell{\textbf{Aesthetic}\\\textbf{Quality}$\uparrow$}
&
\makecell{\textbf{Imaging}\\\textbf{Quality}$\uparrow$}
\\

\midrule

Full-cache
    & 96.94 & 96.15 & 98.08 & \textbf{61.56} & 57.55 & 68.45 \\

\textbf{Conditional}
    & \textbf{97.16} & \textbf{96.29} & \textbf{98.25} & 60.42
    & \textbf{58.27} & \textbf{68.52} \\

\bottomrule
\end{tabular}
\end{center}
\end{table}

\textbf{Response objective.}
Table~\ref{tab:ablation-response} compares full-cache and conditional
responses for both value fitting and merge selection. Conditional responses
improve subject and background consistency, motion smoothness, aesthetic
quality, and imaging quality, with a 0.72-point gain in aesthetic quality
and a 1.14-point reduction in dynamic degree. These results support
block-local response matching as a consolidation objective; qualitative
comparisons appear in Appendix~\ref{app:response-visual}.

\textbf{Sensitivity.}
Appendix~\ref{app:sensitivity} shows that \method retains its quality
advantage across cache budgets and maintains similar scores with moderate
spatial pooling.

\section{Conclusion}

We presented FrameFold, a training-free KV cache consolidation method for autoregressive long-video generation under a fixed budget. By comparing conditional attention responses before and after merging, FrameFold guides both the construction of shared historical representatives and the selection of adjacent merges while preserving complete spatial grids. Experiments on Self Forcing, Causal Forcing, and LongLive demonstrate improved consistency and visual quality under matched cache budgets, with smaller quality declines in continuous eight-minute Self Forcing rollouts. Ablation studies further support conditional-response matching as a consolidation objective and the contributions of representative construction and merge selection.


\subsection*{AI use statement}

In this work, we used generative AI tools to assist with research ideation and
technical discussion, retrieval and discovery of relevant literature and
information, and editing the manuscript for clarity, grammar, and readability.
AI-generated research suggestions were treated as preliminary input and were
independently evaluated by the authors. Relevant literature and factual
information suggested by AI tools were manually verified against the original
sources. AI-assisted manuscript edits were reviewed to ensure that they
preserved the intended technical meaning, experimental claims, and citations.

The proposed method, implementation, experimental results, and scientific
conclusions were ultimately developed, verified, and validated by the authors.
We did not use generative AI tools to generate synthetic datasets or to prove
mathematical claims. We take responsibility for the final content of this work,
including text, claims, code, or artifacts produced with the aid of generative
AI.

\subsection*{Reproducibility statement}
Section~\ref{sec:method} specifies the conditional-response objective,
representative construction, and merge-selection procedure.
Section~\ref{sec:experimental-setup} describes the experimental setup and
evaluation metrics, while Appendix~\ref{app:additional-experiments} provides
study-specific protocols and additional results.
Appendix~\ref{app:implementation-details} details the pooling scheme,
batched candidate evaluation, and baseline configurations. Implementation
code is not included with this submission. We plan to release the source
code publicly, subject to the required internal approvals.

\bibliography{iclr2027_conference}

\begin{thebibliography}{52}
\providecommand{\natexlab}[1]{#1}
\providecommand{\url}[1]{\texttt{#1}}
\expandafter\ifx\csname urlstyle\endcsname\relax
  \providecommand{\doi}[1]{doi: #1}\else
  \providecommand{\doi}{doi: \begingroup \urlstyle{rm}\Url}\fi

\bibitem[Bian et~al.(2026)Bian, Xue, Zhang, Zhang, Jin, Li, Zhuang, Li, Huang, Huang, Duan, and Xu]{bian2026echoinfinity}
Yuxuan Bian, Zeyue Xue, Songchun Zhang, Shiyi Zhang, Weiyang Jin, Yaowei Li, Junhao Zhuang, Haoran Li, Jie Huang, Haoyang Huang, Nan Duan, and Qiang Xu.
\newblock {Echo-Infinity}: Learning evolving memory for real-time infinite video generation.
\newblock \emph{arXiv preprint arXiv:2606.04527}, 2026.

\bibitem[Bolya et~al.(2023)Bolya, Fu, Dai, Zhang, Feichtenhofer, and Hoffman]{bolya2023tome}
Daniel Bolya, Cheng-Yang Fu, Xiaoliang Dai, Peizhao Zhang, Christoph Feichtenhofer, and Judy Hoffman.
\newblock Token merging: Your {ViT} but faster.
\newblock In \emph{International Conference on Learning Representations}, 2023.

\bibitem[Cai et~al.(2026)Cai, Zhang, Liu, Lin, Zhang, Mo, Ma, Zheng, Huang, Liu, and Zhang]{cai2026focusedforcing}
Peiliang Cai, Evelyn Zhang, Jiacheng Liu, Hao Lin, Ruiqi Zhang, Weile Mo, Yue Ma, Shikang Zheng, Jiehang Huang, Dongrui Liu, and Linfeng Zhang.
\newblock {Focused Forcing}: Content-aware per-frame {KV} selection for efficient autoregressive video diffusion.
\newblock \emph{arXiv preprint arXiv:2605.18346}, 2026.

\bibitem[Cai et~al.(2025)Cai, Zhang, Gao, Liu, Li, Liu, Lu, Xiong, Dong, Hu, and Xiao]{cai2025pyramidkv}
Zefan Cai, Yichi Zhang, Bofei Gao, Yuliang Liu, Yucheng Li, Tianyu Liu, Keming Lu, Wayne Xiong, Yue Dong, Junjie Hu, and Wen Xiao.
\newblock {PyramidKV}: Dynamic {KV} cache compression based on pyramidal information funneling.
\newblock In \emph{Conference on Language Modeling}, 2025.

\bibitem[Chen et~al.(2024)Chen, Mart\'{\i}~Mons\'{o}, Du, Simchowitz, Tedrake, and Sitzmann]{chen2024diffusionforcing}
Boyuan Chen, Diego Mart\'{\i}~Mons\'{o}, Yilun Du, Max Simchowitz, Russ Tedrake, and Vincent Sitzmann.
\newblock {Diffusion Forcing}: Next-token prediction meets full-sequence diffusion.
\newblock In \emph{Advances in Neural Information Processing Systems}, 2024.

\bibitem[Chen et~al.(2025)Chen, Lin, Yang, Lin, Zhu, Fan, Zhang, Chen, Chen, Ma, Xiong, Wang, Pang, Kang, Xu, Jin, Liang, Song, Zhao, Xu, Qiu, Li, Fei, Li, and Zhou]{chen2025skyreelsv2}
Guibin Chen, Dixuan Lin, Jiangping Yang, Chunze Lin, Junchen Zhu, Mingyuan Fan, Hao Zhang, Sheng Chen, Zheng Chen, Chengcheng Ma, Weiming Xiong, Wei Wang, Nuo Pang, Kang Kang, Zhiheng Xu, Yuzhe Jin, Yupeng Liang, Yubing Song, Peng Zhao, Boyuan Xu, Di~Qiu, Debang Li, Zhengcong Fei, Yang Li, and Yahui Zhou.
\newblock {SkyReels-V2}: Infinite-length film generative model.
\newblock \emph{arXiv preprint arXiv:2504.13074}, 2025.

\bibitem[Chen et~al.(2026{\natexlab{a}})Chen, Xu, Yang, Chen, and Deng]{chen2026pafukv}
Hanmo Chen, Chenghao Xu, Xu~Yang, Xuan Chen, and Cheng Deng.
\newblock Past- and future-informed {KV} cache policy with salience estimation in autoregressive video diffusion.
\newblock \emph{arXiv preprint arXiv:2601.21896}, 2026{\natexlab{a}}.

\bibitem[Chen et~al.(2026{\natexlab{b}})Chen, Wei, Sun, Shen, Nie, Zou, Zhang, Yang, and Chen]{chen2026contextforcing}
Shuo Chen, Cong Wei, Sun Sun, Tiancheng Shen, Ping Nie, Kai Zou, Ge~Zhang, Ming-Hsuan Yang, and Wenhu Chen.
\newblock {Context Forcing}: Consistent autoregressive video generation with long context.
\newblock In \emph{Proceedings of the International Conference on Machine Learning}, 2026{\natexlab{b}}.

\bibitem[Cui et~al.(2026)Cui, Wu, Li, Yang, Li, Wang, Bai, Ban, and Hsieh]{cui2025selfforcingpp}
Jiaxing Cui, Jie Wu, Ming Li, Tao Yang, Xiaojie Li, Rui Wang, Andrew Bai, Yuanhao Ban, and Cho-Jui Hsieh.
\newblock {Self-Forcing++}: Towards minute-scale high-quality video generation.
\newblock In \emph{International Conference on Learning Representations}, 2026.

\bibitem[Gao et~al.(2026)Gao, Wang, Zhu, Chen, Liu, Bai, Wang, Yuan, Wang, Lu, Cheng, Zhang, Gao, Feng, Liu, Yao, Xu, Zhu, Shen, and Ouyang]{lingbot-world-v2}
Zelin Gao, Qiuyu Wang, Jiapeng Zhu, Jingye Chen, Zichen Liu, Qingyan Bai, Jiahao Wang, Yufeng Yuan, Hanlin Wang, Yichong Lu, Ka~Leong Cheng, Haojie Zhang, Jian Gao, Tianrui Feng, Yuzheng Liu, Yao Yao, Yinghao Xu, Xing Zhu, Yujun Shen, and Hao Ouyang.
\newblock Infinite worlds with versatile interactions.
\newblock \emph{arXiv preprint arXiv:2607.07534}, 2026.

\bibitem[Ge et~al.(2024)Ge, Zhang, Liu, Zhang, Han, and Gao]{ge2024fastgen}
Suyu Ge, Yunan Zhang, Liyuan Liu, Minjia Zhang, Jiawei Han, and Jianfeng Gao.
\newblock Model tells you what to discard: Adaptive {KV} cache compression for {LLMs}.
\newblock In \emph{International Conference on Learning Representations}, 2024.

\bibitem[Guo et~al.(2026{\natexlab{a}})Guo, Jia, Li, Li, Cai, Wang, Li, and Lu]{guo2026dummyforcing}
Hang Guo, Zhaoyang Jia, Jiahao Li, Bin Li, Yuanhao Cai, Jiangshan Wang, Yawei Li, and Yan Lu.
\newblock Efficient autoregressive video diffusion with dummy head.
\newblock \emph{arXiv preprint arXiv:2601.20499}, 2026{\natexlab{a}}.

\bibitem[Guo et~al.(2026{\natexlab{b}})Guo, Yang, He, Zhao, Wei, Yang, Huang, and Lin]{guo2025resamplingforcing}
Yuwei Guo, Ceyuan Yang, Hao He, Yang Zhao, Meng Wei, Zhenheng Yang, Weilin Huang, and Dahua Lin.
\newblock End-to-end training for autoregressive video diffusion via self-resampling.
\newblock In \emph{Proceedings of the European Conference on Computer Vision}, 2026{\natexlab{b}}.

\bibitem[Hu et~al.(2026)Hu, Yang, Huang, Han, and Chen]{hu2026longliverag}
Qixin Hu, Shuai Yang, Wei Huang, Song Han, and Yukang Chen.
\newblock {LongLive-RAG}: A general retrieval-augmented framework for long video generation.
\newblock \emph{arXiv preprint arXiv:2606.02553}, 2026.

\bibitem[Huang et~al.(2026{\natexlab{a}})Huang, Wu, Zhou, Miao, and Long]{huang2026vid2world}
Siqiao Huang, Jialong Wu, Qixing Zhou, Shangchen Miao, and Mingsheng Long.
\newblock {Vid2World}: Crafting video diffusion models to interactive world models.
\newblock In \emph{International Conference on Learning Representations}, 2026{\natexlab{a}}.

\bibitem[Huang et~al.(2025)Huang, Li, He, Zhou, and Shechtman]{huang2025selfforcing}
Xun Huang, Zhengqi Li, Guande He, Mingyuan Zhou, and Eli Shechtman.
\newblock {Self Forcing}: Bridging the train-test gap in autoregressive video diffusion.
\newblock In \emph{Advances in Neural Information Processing Systems}, 2025.

\bibitem[Huang et~al.(2024)Huang, He, Yu, Zhang, Si, Jiang, Zhang, Wu, Jin, Chanpaisit, Wang, Chen, Wang, Lin, Qiao, and Liu]{huang2024vbench}
Ziqi Huang, Yinan He, Jiashuo Yu, Fan Zhang, Chenyang Si, Yuming Jiang, Yuanhan Zhang, Tianxing Wu, Qingyang Jin, Nattapol Chanpaisit, Yaohui Wang, Xinyuan Chen, Limin Wang, Dahua Lin, Yu~Qiao, and Ziwei Liu.
\newblock {VBench}: Comprehensive benchmark suite for video generative models.
\newblock In \emph{Proceedings of the IEEE/CVF Conference on Computer Vision and Pattern Recognition}, 2024.

\bibitem[Huang et~al.(2026{\natexlab{b}})Huang, Zhang, Xu, He, Yu, Dong, Ma, Chanpaisit, Si, Jiang, Wang, Chen, Chen, Wang, Lin, Qiao, and Liu]{huang2026vbenchpp}
Ziqi Huang, Fan Zhang, Xiaojie Xu, Yinan He, Jiashuo Yu, Ziyue Dong, Qianli Ma, Nattapol Chanpaisit, Chenyang Si, Yuming Jiang, Yaohui Wang, Xinyuan Chen, Ying-Cong Chen, Limin Wang, Dahua Lin, Yu~Qiao, and Ziwei Liu.
\newblock {VBench++}: Comprehensive and versatile benchmark suite for video generative models.
\newblock \emph{IEEE Transactions on Pattern Analysis and Machine Intelligence}, 2026{\natexlab{b}}.

\bibitem[Ji et~al.(2026)Ji, Zhong, Zhang, Yang, Jin, Qin, Luo, Mao, Liu, and Li]{ji2026forcingkv}
Yicheng Ji, Zhizhou Zhong, Jun Zhang, Qin Yang, XiTai Jin, Ying Qin, Wenhan Luo, Shuiyang Mao, Wei Liu, and Huan Li.
\newblock {Forcing-KV}: Hybrid {KV} cache compression for efficient autoregressive video diffusion models.
\newblock \emph{arXiv preprint arXiv:2605.09681}, 2026.

\bibitem[Kim et~al.(2026)Kim, Hu, Kuo, and Beerel]{kim2026memrope}
Youngrae Kim, Qixin Hu, C.-C.~Jay Kuo, and Peter Beerel.
\newblock {MemRoPE}: Training-free infinite video generation via evolving memory tokens.
\newblock In \emph{Proceedings of the European Conference on Computer Vision}, 2026.

\bibitem[Li et~al.(2026{\natexlab{a}})Li, Liu, Lin, and Chandraker]{li2026rollingsink}
Haodong Li, Shaoteng Liu, Zhe Lin, and Manmohan Chandraker.
\newblock {Rolling Sink}: Bridging limited-horizon training and open-ended testing in autoregressive video diffusion.
\newblock \emph{arXiv preprint arXiv:2602.07775}, 2026{\natexlab{a}}.

\bibitem[Li et~al.(2026{\natexlab{b}})Li, Pan, Luan, Gao, and Alahi]{li2025stablevideoinfinity}
Wuyang Li, Wentao Pan, Po-Chien Luan, Yang Gao, and Alexandre Alahi.
\newblock {Stable Video Infinity}: Infinite-length video generation with error recycling.
\newblock In \emph{International Conference on Learning Representations}, 2026{\natexlab{b}}.

\bibitem[Li et~al.(2024{\natexlab{a}})Li, Ma, Yang, and Yang]{li2023vidtome}
Xirui Li, Chao Ma, Xiaokang Yang, and Ming-Hsuan Yang.
\newblock {VidToMe}: Video token merging for zero-shot video editing.
\newblock In \emph{Proceedings of the IEEE/CVF Conference on Computer Vision and Pattern Recognition}, 2024{\natexlab{a}}.

\bibitem[Li et~al.(2024{\natexlab{b}})Li, Huang, Yang, Venkitesh, Locatelli, Ye, Cai, Lewis, and Chen]{li2024snapkv}
Yuhong Li, Yingbing Huang, Bowen Yang, Bharat Venkitesh, Acyr Locatelli, Hanchen Ye, Tianle Cai, Patrick Lewis, and Deming Chen.
\newblock {SnapKV}: {LLM} knows what you are looking for before generation.
\newblock In \emph{Advances in Neural Information Processing Systems}, 2024{\natexlab{b}}.

\bibitem[Liu et~al.(2024)Liu, Liu, Pan, He, Haffari, and Zhuang]{liu2024minicache}
Akide Liu, Jing Liu, Zizheng Pan, Yefei He, Gholamreza Haffari, and Bohan Zhuang.
\newblock {MiniCache}: {KV} cache compression in depth dimension for large language models.
\newblock In \emph{Advances in Neural Information Processing Systems}, 2024.

\bibitem[Liu et~al.(2026)Liu, Hu, Xu, Shan, and Lu]{liu2025rollingforcing}
Kunhao Liu, Wenbo Hu, Jiale Xu, Ying Shan, and Shijian Lu.
\newblock {Rolling Forcing}: Autoregressive long video diffusion in real time.
\newblock In \emph{International Conference on Learning Representations}, 2026.

\bibitem[Liu et~al.(2023)Liu, Desai, Liao, Wang, Xie, Xu, Kyrillidis, and Shrivastava]{liu2023scissorhands}
Zichang Liu, Aditya Desai, Fangshuo Liao, Weitao Wang, Victor Xie, Zhaozhuo Xu, Anastasios Kyrillidis, and Anshumali Shrivastava.
\newblock {Scissorhands}: Exploiting the persistence of importance hypothesis for {LLM KV} cache compression at test time.
\newblock In \emph{Advances in Neural Information Processing Systems}, 2023.

\bibitem[Lu et~al.(2026)Lu, Yang, Koniusz, Song, and Yang]{lu2026fademem}
Yu~Lu, Junjie Yang, Piotr Koniusz, YuXin Song, and Yi~Yang.
\newblock {FadeMem}: Distance-aware memory consolidation for autoregressive video diffusion.
\newblock \emph{arXiv preprint arXiv:2606.10671}, 2026.

\bibitem[Luo et~al.(2026)Luo, Liu, Wang, Liu, Chen, Wang, Zhu, Gao, Hu, Sun, and Chen]{luo2026futureforcing}
Jiayi Luo, Qiyan Liu, Tengyang Wang, JunHao Liu, Jiayu Chen, Cong Wang, Hanxin Zhu, Chen Gao, Xiaobin Hu, Qingyun Sun, and Zhibo Chen.
\newblock {Future Forcing}: Future-aware training-free {KV} cache policy for autoregressive video generation.
\newblock \emph{arXiv preprint arXiv:2605.30083}, 2026.

\bibitem[Ma et~al.(2026)Ma, Zheng, Xu, Xu, Ling, Zheng, Kuang, Li, Wang, Xiao, Chao, and Ji]{ma2026flowcache}
Yuexiao Ma, Xuzhe Zheng, Jing Xu, Xiwei Xu, Feng Ling, Xiawu Zheng, Huafeng Kuang, Huixia Li, Xing Wang, Xuefeng Xiao, Fei Chao, and Rongrong Ji.
\newblock Flow caching for autoregressive video generation.
\newblock In \emph{International Conference on Learning Representations}, 2026.

\bibitem[Mao et~al.(2025)Mao, Li, Li, Xu, Ying, He, Pang, Qiao, and Zhang]{mao2025yume15}
Xiaofeng Mao, Zhen Li, Chuanhao Li, Xiaojie Xu, Kaining Ying, Tong He, Jiangmiao Pang, Yu~Qiao, and Kaipeng Zhang.
\newblock {Yume-1.5}: A text-controlled interactive world generation model.
\newblock \emph{arXiv preprint arXiv:2512.22096}, 2025.

\bibitem[Meng et~al.(2026)Meng, Luo, Li, Jiang, Gao, Chen, Li, and Zhang]{meng2026tethercache}
Yu~Meng, Xiangyang Luo, Letian Li, Wenyuan Jiang, Chen Gao, Xinlei Chen, Yong Li, and Xiao-Ping Zhang.
\newblock {TetherCache}: Stabilizing autoregressive long-form video generation with gated recall and trusted alignment.
\newblock \emph{arXiv preprint arXiv:2606.13035}, 2026.

\bibitem[Polyak et~al.(2024)Polyak, Zohar, Brown, Tjandra, Sinha, Lee, Vyas, Shi, Ma, Chuang, Yan, Choudhary, Wang, Sethi, Pang, Ma, Misra, Hou, Wang, Jagadeesh, Li, Zhang, Singh, Williamson, Le, Yu, Singh, Zhang, Vajda, Duval, Girdhar, Sumbaly, Rambhatla, Tsai, Azadi, Datta, Chen, Bell, Ramaswamy, Sheynin, Bhattacharya, Motwani, Xu, Li, Hou, Hsu, Yin, Dai, Taigman, Luo, Liu, Wu, Zhao, Kirstain, He, He, Pumarola, Thabet, Sanakoyeu, Mallya, Guo, Araya, Kerr, Wood, Liu, Peng, Vengertsev, Schonfeld, Blanchard, Juefei-Xu, Nord, Liang, Hoffman, Kohler, Fire, Sivakumar, Chen, Yu, Gao, Georgopoulos, Moritz, Sampson, Li, Parmeggiani, Fine, Fowler, Petrovic, and Du]{polyak2024moviegen}
Adam Polyak, Amit Zohar, Andrew Brown, Andros Tjandra, Animesh Sinha, Ann Lee, Apoorv Vyas, Bowen Shi, Chih-Yao Ma, Ching-Yao Chuang, David Yan, Dhruv Choudhary, Dingkang Wang, Geet Sethi, Guan Pang, Haoyu Ma, Ishan Misra, Ji~Hou, Jialiang Wang, Kiran Jagadeesh, Kunpeng Li, Luxin Zhang, Mannat Singh, Mary Williamson, Matt Le, Matthew Yu, Mitesh~Kumar Singh, Peizhao Zhang, Peter Vajda, Quentin Duval, Rohit Girdhar, Roshan Sumbaly, Sai~Saketh Rambhatla, Sam Tsai, Samaneh Azadi, Samyak Datta, Sanyuan Chen, Sean Bell, Sharadh Ramaswamy, Shelly Sheynin, Siddharth Bhattacharya, Simran Motwani, Tao Xu, Tianhe Li, Tingbo Hou, Wei-Ning Hsu, Xi~Yin, Xiaoliang Dai, Yaniv Taigman, Yaqiao Luo, Yen-Cheng Liu, Yi-Chiao Wu, Yue Zhao, Yuval Kirstain, Zecheng He, Zijian He, Albert Pumarola, Ali Thabet, Artsiom Sanakoyeu, Arun Mallya, Baishan Guo, Boris Araya, Breena Kerr, Carleigh Wood, Ce~Liu, Cen Peng, Dimitry Vengertsev, Edgar Schonfeld, Elliot Blanchard, Felix Juefei-Xu, Fraylie Nord, Jeff Liang, John Hoffman, Jonas
  Kohler, Kaolin Fire, Karthik Sivakumar, Lawrence Chen, Licheng Yu, Luya Gao, Markos Georgopoulos, Rashel Moritz, Sara~K. Sampson, Shikai Li, Simone Parmeggiani, Steve Fine, Tara Fowler, Vladan Petrovic, and Yuming Du.
\newblock {Movie Gen}: A cast of media foundation models.
\newblock \emph{arXiv preprint arXiv:2410.13720}, 2024.

\bibitem[Qian et~al.(2026)Qian, Wang, Zhang, Zou, Yu, Li, Liu, Li, Kang, Huang, An, Zhang, Jiang, Wang, Sun, Liu, and Li]{qian2026matrixgame35}
Runjia Qian, Zile Wang, Jihai Zhang, Kai Zou, Wei Yu, Jiaxing Li, Zexiang Liu, Yaokun Li, Fei Kang, Kaichen Huang, Mengyin An, Haobo Zhang, Biao Jiang, Jiahua Wang, Haofeng Sun, Yang Liu, and Yangguang Li.
\newblock {Matrix-Game 3.5}: Enhancing real-time streaming interactive world models with patch memory.
\newblock \emph{arXiv preprint arXiv:2608.29910}, 2026.

\bibitem[Samuel et~al.(2026)Samuel, Tzachor, Levy, Green, Chechik, and Ben-Ari]{samuel2026fastar}
Dvir Samuel, Issar Tzachor, Matan Levy, Michael Green, Gal Chechik, and Rami Ben-Ari.
\newblock {FAST-AR}: Fast autoregressive video diffusion and world models with temporal cache compression and sparse attention.
\newblock In \emph{Proceedings of the International Conference on Machine Learning}, 2026.

\bibitem[{Sand.ai} et~al.(2025){Sand.ai}, Teng, Jia, Sun, Li, Li, Tang, Han, Zhang, Zhang, Luo, Kang, Sun, Cao, Huang, Lin, Fang, Tao, Zhang, Wang, Liu, Shi, Su, Sun, Pan, Wang, Sheng, Cui, Hu, Yan, Yin, Zhang, Liu, Yin, Yang, Song, Hu, Zhang, and Li]{sandai2025magi1}
{Sand.ai}, Hansi Teng, Hongyu Jia, Lei Sun, Lingzhi Li, Maolin Li, Mingqiu Tang, Shuai Han, Tianning Zhang, W.~Q. Zhang, Weifeng Luo, Xiaoyang Kang, Yuchen Sun, Yue Cao, Yunpeng Huang, Yutong Lin, Yuxin Fang, Zewei Tao, Zheng Zhang, Zhongshu Wang, Zixun Liu, Dai Shi, Guoli Su, Hanwen Sun, Hong Pan, Jie Wang, Jiexin Sheng, Min Cui, Min Hu, Ming Yan, Shucheng Yin, Siran Zhang, Tingting Liu, Xianping Yin, Xiaoyu Yang, Xin Song, Xuan Hu, Yankai Zhang, and Yuqiao Li.
\newblock {MAGI-1}: Autoregressive video generation at scale.
\newblock \emph{arXiv preprint arXiv:2505.13211}, 2025.

\bibitem[Su et~al.(2024)Su, Ahmed, Lu, Pan, Bo, and Liu]{su2021roformer}
Jianlin Su, Murtadha Ahmed, Yu~Lu, Shengfeng Pan, Wen Bo, and Yunfeng Liu.
\newblock {RoFormer}: Enhanced transformer with rotary position embedding.
\newblock \emph{Neurocomputing}, 2024.

\bibitem[Sun et~al.(2026)Sun, Zhang, Wang, Wu, Wang, Wang, Wang, Zhang, Wang, and Guo]{sun2026worldplay}
Wenqiang Sun, Haiyu Zhang, Haoyuan Wang, Junta Wu, Zehan Wang, Zhenwei Wang, Yunhong Wang, Jun Zhang, Tengfei Wang, and Chunchao Guo.
\newblock {WorldPlay}: Towards long-term geometric consistency for real-time interactive world modeling.
\newblock In \emph{Proceedings of the International Conference on Machine Learning}, 2026.

\bibitem[Wang et~al.(2024)Wang, Jin, Yu, and Zhang]{wang2024kvmerger}
Zheng Wang, Boxiao Jin, Zhongzhi Yu, and Minjia Zhang.
\newblock Model tells you where to merge: Adaptive {KV} cache merging for {LLMs} on long-context tasks.
\newblock \emph{arXiv preprint arXiv:2407.08454}, 2024.

\bibitem[Xi et~al.(2026)Xi, Yang, Zhao, Li, Cai, Li, Lin, Zhang, Zhang, Li, Xu, Wu, Xu, Stoica, Han, and Keutzer]{xi2026quantvideogen}
Haocheng Xi, Shuo Yang, Yilong Zhao, Muyang Li, Han Cai, Xingyang Li, Yujun Lin, Zhuoyang Zhang, Jintao Zhang, Xiuyu Li, Zhiying Xu, Jun Wu, Chenfeng Xu, Ion Stoica, Song Han, and Kurt Keutzer.
\newblock {Quant VideoGen}: Auto-regressive long video generation via 2-bit {KV}-cache quantization.
\newblock In \emph{Proceedings of the International Conference on Machine Learning}, 2026.

\bibitem[Xiao et~al.(2024)Xiao, Tian, Chen, Han, and Lewis]{xiao2024streamingllm}
Guangxuan Xiao, Yuandong Tian, Beidi Chen, Song Han, and Mike Lewis.
\newblock Efficient streaming language models with attention sinks.
\newblock In \emph{International Conference on Learning Representations}, 2024.

\bibitem[Yang et~al.(2026)Yang, Huang, Chu, Xiao, Zhao, Wang, Li, Xie, Chen, Lu, Han, and Chen]{yang2025longlive}
Shuai Yang, Wei Huang, Ruihang Chu, Yicheng Xiao, Yuyang Zhao, Xianbang Wang, Muyang Li, Enze Xie, Yingcong Chen, Yao Lu, Song Han, and Yukang Chen.
\newblock {LongLive}: Real-time interactive long video generation.
\newblock In \emph{International Conference on Learning Representations}, 2026.

\bibitem[Yi et~al.(2026)Yi, Jang, Cho, Nam, Yoon, and Kim]{yi2025deepforcing}
Jung Yi, Wooseok Jang, Paul~Hyunbin Cho, Jisu Nam, Heeji Yoon, and Seungryong Kim.
\newblock {Deep Forcing}: Training-free long video generation with deep sink and participative compression.
\newblock In \emph{Proceedings of the International Conference on Machine Learning}, 2026.

\bibitem[Yin et~al.(2025)Yin, Zhang, Zhang, Freeman, Durand, Shechtman, and Huang]{yin2025causvid}
Tianwei Yin, Qiang Zhang, Richard Zhang, William~T. Freeman, Fredo Durand, Eli Shechtman, and Xun Huang.
\newblock From slow bidirectional to fast autoregressive video diffusion models.
\newblock In \emph{Proceedings of the IEEE/CVF Conference on Computer Vision and Pattern Recognition}, 2025.

\bibitem[Zhang et~al.(2025)Zhang, Cai, Li, Wetzstein, and Agrawala]{zhang2025framepack}
Lvmin Zhang, Shengqu Cai, Muyang Li, Gordon Wetzstein, and Maneesh Agrawala.
\newblock Frame context packing and drift prevention in next-frame-prediction video diffusion models.
\newblock In \emph{Advances in Neural Information Processing Systems}, 2025.

\bibitem[Zhang et~al.(2026)Zhang, Cai, Li, Zeng, Lu, Rao, Han, Wetzstein, and Agrawala]{zhang2025tinyhistory}
Lvmin Zhang, Shengqu Cai, Muyang Li, Chong Zeng, Beijia Lu, Anyi Rao, Song Han, Gordon Wetzstein, and Maneesh Agrawala.
\newblock {TinyHistory}: Lightweight video history embeddings via two-stage context learning.
\newblock In \emph{Proceedings of the European Conference on Computer Vision}, 2026.

\bibitem[Zhang et~al.(2023)Zhang, Sheng, Zhou, Chen, Zheng, Cai, Song, Tian, R{\'e}, Barrett, Wang, and Chen]{zhang2023h2o}
Zhenyu Zhang, Ying Sheng, Tianyi Zhou, Tianlong Chen, Lianmin Zheng, Ruisi Cai, Zhao Song, Yuandong Tian, Christopher R{\'e}, Clark Barrett, Zhangyang~``Atlas'' Wang, and Beidi Chen.
\newblock {H2O}: Heavy-hitter oracle for efficient generative inference of large language models.
\newblock In \emph{Advances in Neural Information Processing Systems}, 2023.

\bibitem[Zhao et~al.(2026{\natexlab{a}})Zhao, Chi, Zhang, Ma, Li, Fang, Tang, Gao, and Wu]{zhao2026densitykv}
Wenqu Zhao, Xuemin Chi, Xin Zhang, Guoqing Ma, Baorun Li, Jianjie Fang, Peizhi Tang, Chen Gao, and Wei Wu.
\newblock {DensityKV}: Density-guided {KV} cache compression for long video generation.
\newblock \emph{arXiv preprint arXiv:2608.27922}, 2026{\natexlab{a}}.

\bibitem[Zhao et~al.(2026{\natexlab{b}})Zhao, Lu, Liu, Song, Deng, and Patras]{zhao2026relaxforcing}
Zengqun Zhao, Yanzuo Lu, Ziquan Liu, Jifei Song, Jiankang Deng, and Ioannis Patras.
\newblock {Relax Forcing}: Relaxed {KV}-memory for consistent long video generation.
\newblock In \emph{Proceedings of the British Machine Vision Conference}, 2026{\natexlab{b}}.
\newblock Accepted.

\bibitem[Zhu et~al.(2026)Zhu, Zhao, He, Su, Li, and Zhu]{zhu2026causalforcing}
Hongzhou Zhu, Min Zhao, Guande He, Hang Su, Chongxuan Li, and Jun Zhu.
\newblock {Causal Forcing}: Autoregressive diffusion distillation done right for high-quality real-time interactive video generation.
\newblock In \emph{Proceedings of the International Conference on Machine Learning}, 2026.

\bibitem[Zhu et~al.(2025)Zhu, Zhang, Sun, Tian, and Tang]{zhu2025memorizeandgenerate}
Tianrui Zhu, Shiyi Zhang, Zhirui Sun, Jingqi Tian, and Yansong Tang.
\newblock {Memorize-and-Generate}: Towards long-term consistency in real-time video generation.
\newblock \emph{arXiv preprint arXiv:2512.18741}, 2025.

\bibitem[Zweiger et~al.(2026)Zweiger, Fu, Guo, and Kim]{zweiger2026attentionmatching}
Adam Zweiger, Xinghong Fu, Han Guo, and Yoon Kim.
\newblock Fast {KV} compaction via {Attention Matching}.
\newblock In \emph{Proceedings of the International Conference on Machine Learning}, 2026.

\end{thebibliography}
\bibliographystyle{iclr2027_conference}

\clearpage
\appendix
\addtocontents{toc}{\protect\setcounter{tocdepth}{2}}
\begingroup
    \renewcommand{\contentsname}{Appendix Contents}
    \setcounter{tocdepth}{-1}
    \hypersetup{linktoc=all}
    \pdfbookmark[0]{Appendix Contents}{appendix-contents}
    \tableofcontents
\endgroup
\setcounter{tocdepth}{2}
\clearpage

\section{Supplementary Experiments and Analyses}
\label{app:additional-experiments}

\subsection{Motivation Analysis: Spatial Attention and Token Selection}
\label{app:selection-maps}

\paragraph{Analysis objective.}
Queries at different spatial locations share the same retained history,
so a shared token selection must serve their different retrieval needs.
We supplement the analysis of shared ranking in
Section~\ref{sec:attention-retention-limits} through query-specific
attention distributions and the spatial patterns produced by actual
Top-K selection.

\begin{figure}[!htb]
    \centering
    \includegraphics[width=\linewidth]{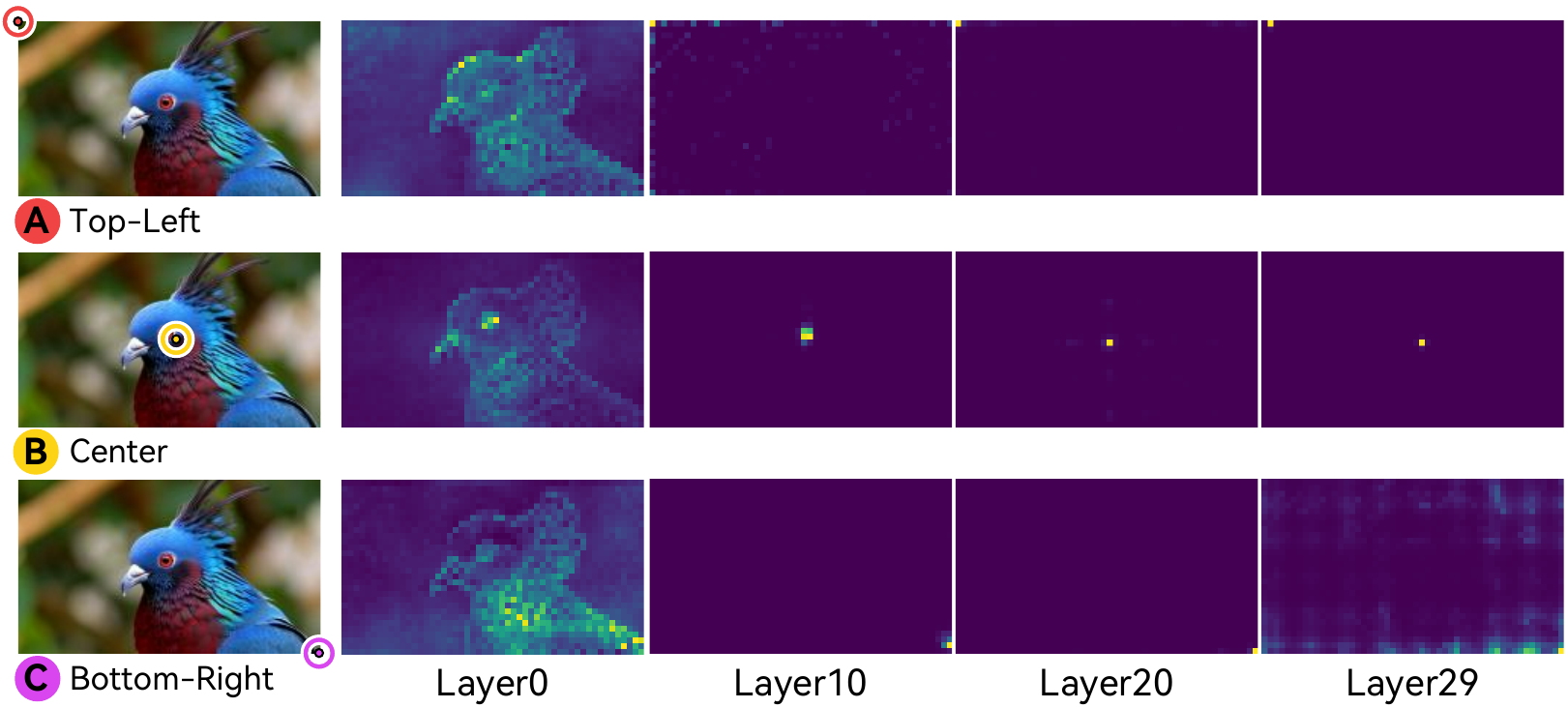}
    \caption{\textbf{Visualization of query-specific spatial attention patterns.}
    Each row shows a query at the upper-left, center, or lower-right
    spatial location, followed by its attention maps in Transformer
    layers 0, 10, 20, and 29.}
    \label{fig:query-spatial-attention}
\end{figure}

\paragraph{Query-specific attention patterns.}
Figure~\ref{fig:query-spatial-attention} shows attention distributions for
queries at the upper-left, center, and lower-right locations in layers
0, 10, 20, and 29. Within the same layer, regions with high attention
differ across queries: in layers 10 and 20, the hotspots are concentrated
near the respective query locations. The distributions also vary across
layers. For example, attention is more dispersed in layer 0, and the
lower-right query also exhibits a broader response pattern in layer 29.
These observations illustrate differences in the historical content
attended to by spatial queries; the regions emphasized by one query do
not represent the needs of the others.

\begin{figure}[!htb]
    \centering
    \includegraphics[width=\linewidth]{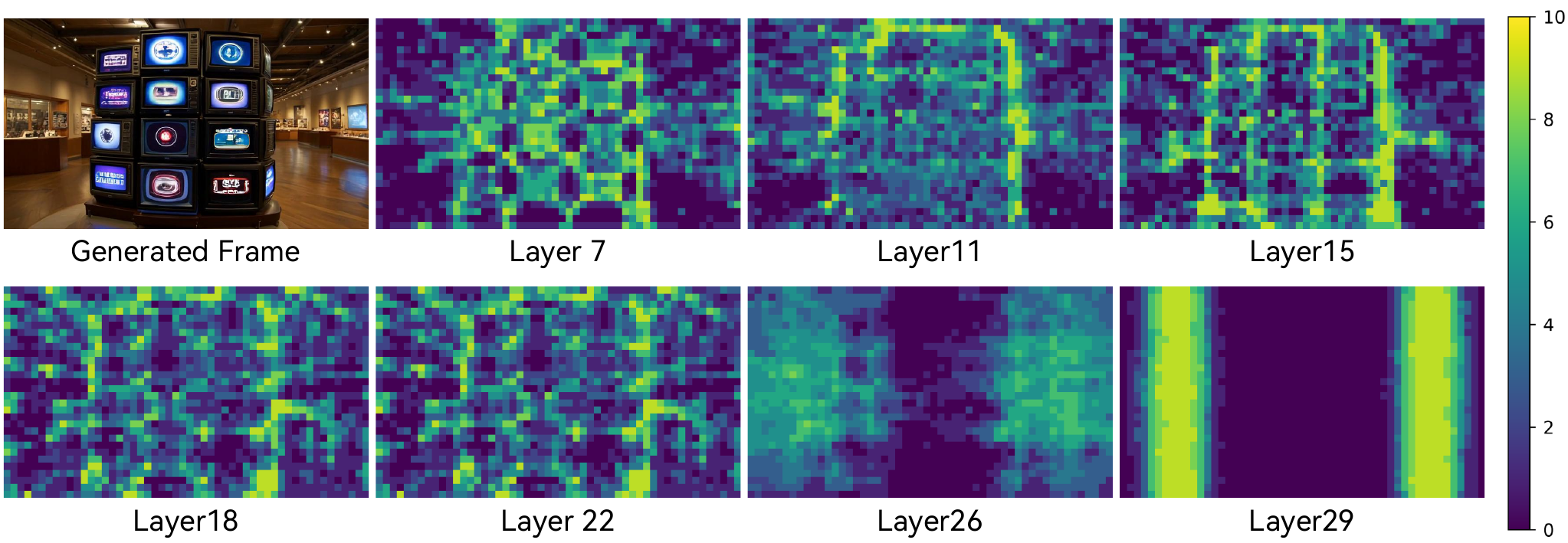}
    \caption{\textbf{Visualization of spatial selection patterns in Deep Forcing.}
    A generated frame is shown alongside selection-frequency maps for
    seven layers: 7, 11, 15, 18, 22, 26, and 29. Color indicates selection
    frequency at each spatial location.}
    \label{fig:topk-selection-maps}
\end{figure}

\paragraph{Spatial selection patterns.}
We further examine actual Top-K selection in Deep Forcing.
Figure~\ref{fig:topk-selection-maps} shows selection-frequency maps from
seven layers. Layers 7, 11, 15, 18, and 22 exhibit scattered local
concentrations and bands of high selection frequency. In layer 26,
selection frequencies are higher on the left and right sides and lower
in the center. Layer 29 exhibits two prominent vertical bands of high
selection frequency, with a large central region close to zero. These
maps illustrate the spatial retention patterns produced by shared Top-K
ranking: selection frequencies are uneven across spatial locations, and
the distributions vary across layers. Together with the query-level
coverage gaps in Section~\ref{sec:attention-retention-limits}, these
observations motivate accounting for retrieval needs at different spatial
locations during history compression and preserving complete spatial
representations.

\subsection{Motivation Analysis: Attention Coverage and Generation Quality}
\label{app:topk-bottomk-visual}

\paragraph{Experimental setup.}
This section supplements the Top-K/Bottom-K comparison in
Section~\ref{sec:attention-retention-limits} with key experimental settings
and temporal results. We generate 60-second videos on 48 matched prompts
using Self Forcing and Causal Forcing. Both policies retain 50\% of the
candidate history tokens and share the same attention-scoring procedure,
prompts, random seeds, and sampling settings; they retain the
highest-ranked and lowest-ranked candidates, respectively. For the
attention coverage in Figure~\ref{fig:motivation}(a), we measure the
attention assigned by each query to retained tokens as a fraction of its
total attention within the candidate history.

\begin{figure}[!htb]
    \centering
    \includegraphics[width=0.97\linewidth]{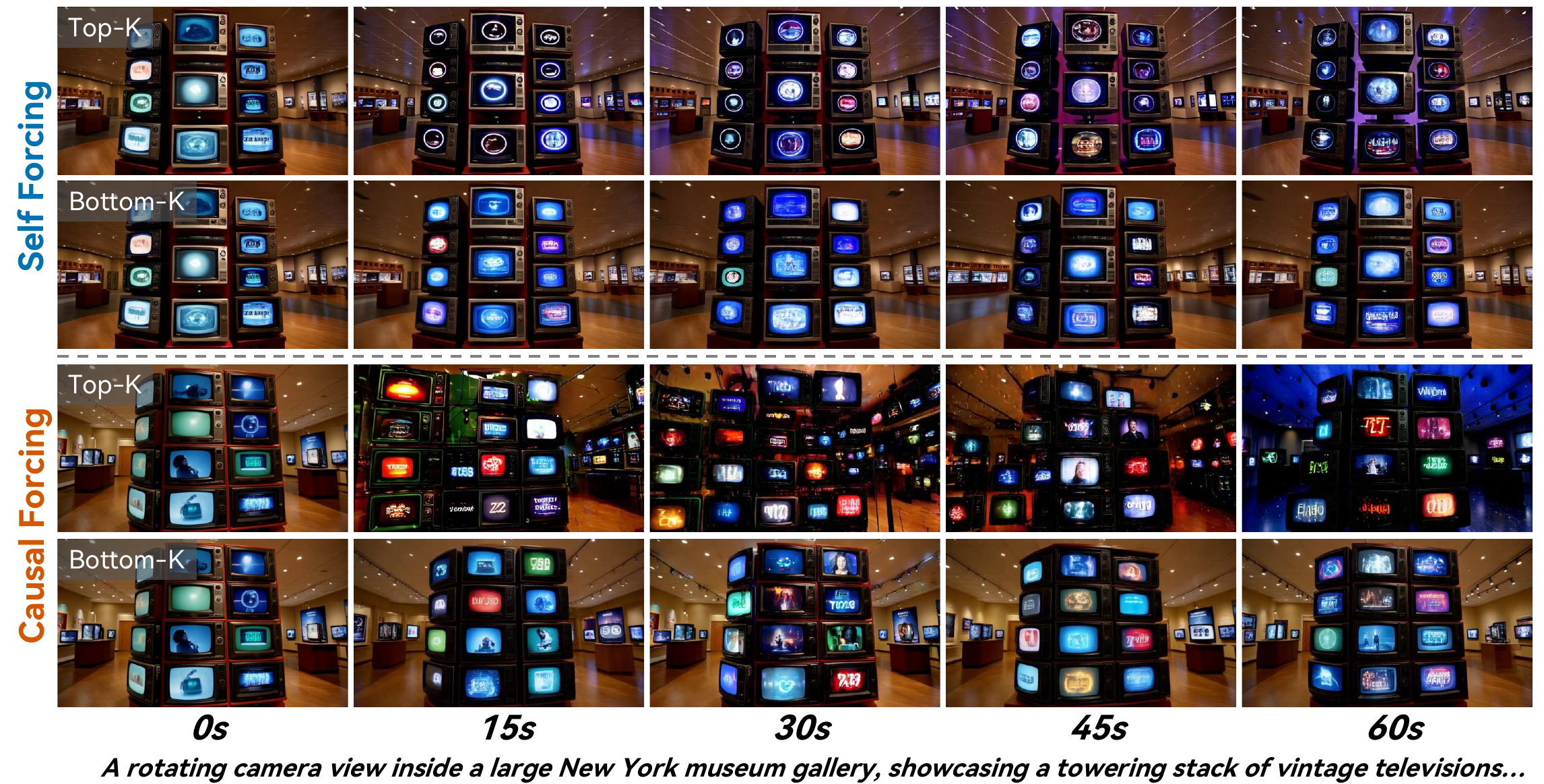}
    \caption{\textbf{Qualitative comparison of Top-K and Bottom-K over
    60 seconds of generation.}
    Results are shown for Self Forcing and Causal Forcing. Both policies
    retain 50\% of the candidate history using the same scoring procedure,
    with matched prompts, seeds, and sampling settings.}
    \label{fig:topk-bottomk-temporal}
\end{figure}

\paragraph{Temporal results.}
Figure~\ref{fig:topk-bottomk-temporal} extends the first- and final-frame
comparison in Figure~\ref{fig:motivation}(c), showing results at 0, 15,
30, 45, and 60 seconds. In the Causal Forcing gallery example, the
intermediate frames at 15--45 seconds under Top-K already exhibit lower
overall brightness and changes in background color, followed by a
pronounced blue cast in the final 60-second frame. Bottom-K better
preserves the warm tones of the initial gallery background across the
sampled timestamps. These intermediate frames show how appearance
differences between the two retention policies develop over time,
providing a visual complement to the observation in the main text that
higher current attention coverage need not yield better subsequent
generation quality.

\subsection{Supplementary Results: Qualitative Comparisons}
\label{app:qualitative-comparisons}

\begin{figure}[!htb]
    \centering
    \includegraphics[width=\linewidth]{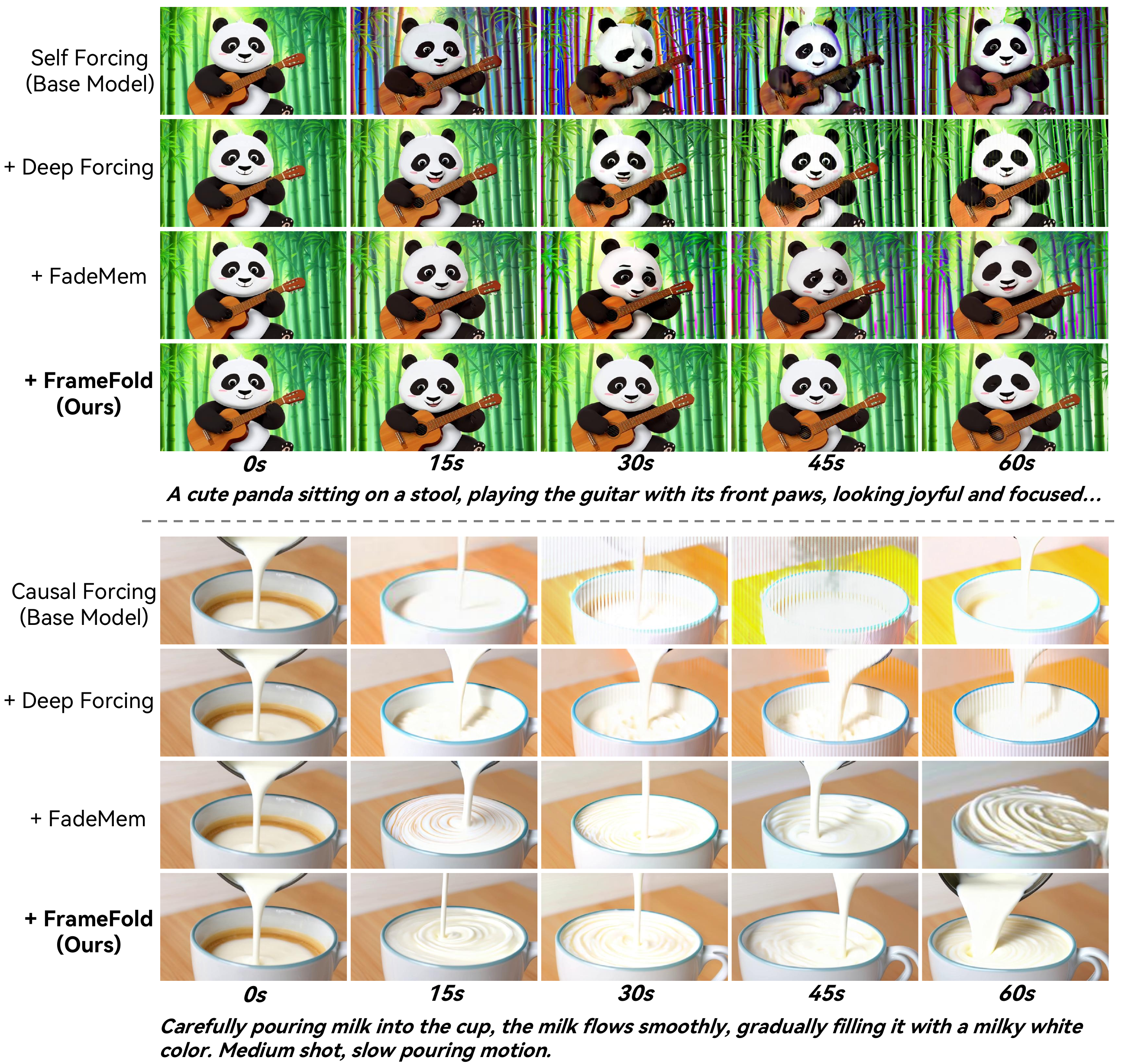}
    \caption{\textbf{Additional qualitative comparisons on 60-second
    video generation.}
    Self Forcing and Causal Forcing examples are shown in the top and
    bottom groups, respectively, with frames sampled at 0, 15, 30, 45,
    and 60 seconds. Methods within each example use the same prompt,
    seed, and sampling settings, with a cache budget of 12
    frame-equivalent KV entries.}
    \label{fig:additional-qualitative}
\end{figure}

Figure~\ref{fig:additional-qualitative} supplements the main comparison in
Section~\ref{sec:main-results} with two additional examples under the same
60-second generation settings. The Self Forcing example shows a panda
playing a guitar, and the Causal Forcing example shows milk being poured
into a cup.

In the panda example, the original Self Forcing model develops colored
stripe artifacts in the bamboo background, while FadeMem exhibits changes
in the panda's facial appearance and background color. \method maintains
a more consistent panda appearance and more stable background colors.
In the milk-pouring example, the original Causal Forcing model and
Deep Forcing exhibit vertical stripe artifacts in later frames, whereas
\method maintains cleaner cup contours and background textures.

\subsection{Supplementary Results: Long-Horizon Generation}
\label{app:long-horizon}

\paragraph{Evaluation protocol.}
This section supplements the long-horizon experiment in
Section~\ref{sec:long-horizon} with complete results across all six
VBench-Long dimensions and temporal visualizations. We compare \method,
FadeMem, and Deep Forcing using Self Forcing on 24 prompts, with a fixed
cache budget of 12 frame-equivalent KV entries. Each method generates a
continuous 480-second video for each prompt. We evaluate its nested
prefixes at 30, 60, 120, 240, and 480 seconds to track generation
performance as the evaluated duration increases.

\paragraph{Quantitative results.}
Table~\ref{tab:ab-duration} shows that \method achieves the highest subject
consistency, background consistency, motion smoothness, aesthetic quality,
and imaging quality among the three methods at every evaluated duration,
with smaller declines from 30 to 480 seconds. For example, imaging quality
decreases by 1.02 points for \method, compared with 2.09 points for FadeMem
and 3.64 points for Deep Forcing. Over the same interval, \method's motion
smoothness changes from 98.62 to 98.56, while its dynamic degree increases
from 67.45 to 69.79. Its consistency and visual-quality advantages thus
accompany high motion smoothness and dynamic degree.

\begin{table}[htbp]
\caption{
\textbf{Quantitative comparison of \method with baselines on video
generation up to 480 seconds.}
We evaluate Self Forcing on 24 prompts with a cache budget of 12
frame-equivalent KV entries. At each duration, we evaluate a nested prefix of the same
continuous 480-second rollout. Higher is better; best scores within each
duration group are in \textbf{bold}.
}
\label{tab:ab-duration}

\setlength{\tabcolsep}{2.5pt}

\begin{center}
\begin{tabular}{
    c
    l
    c c c c c c 
}
\toprule

\textbf{Duration}
&
\textbf{Method}
&
\makecell{\textbf{Subject}\\\textbf{Consist.}$\uparrow$}
&
\makecell{\textbf{Background}\\\textbf{Consist.}$\uparrow$}
&
\makecell{\textbf{Motion}\\\textbf{Smooth.}$\uparrow$}
&
\makecell{\textbf{Dynamic}\\\textbf{Degree}$\uparrow$}
&
\makecell{\textbf{Aesthetic}\\\textbf{Quality}$\uparrow$}
&
\makecell{\textbf{Imaging}\\\textbf{Quality}$\uparrow$}
\\

\midrule


\multirow[c]{3}{*}{30s}
&
Deep Forcing
    & 97.27 & 96.26 & 98.49 & 55.99 
    & 61.40 & 69.50 \\

&
FadeMem
    & 97.22 & 96.29 & 98.50 & 67.44
    & 61.77 & 71.77  \\

&
\ffcell{\textbf{FrameFold}}
    & \ffcell{\textbf{97.36}}
    & \ffcell{\textbf{96.35}}
    & \ffcell{\textbf{98.62}}
    & \ffcell{\textbf{67.45}}
    & \ffcell{\textbf{62.00}}
    & \ffcell{\textbf{72.31}}
\\

\midrule


\multirow[c]{3}{*}{60s}
&
Deep Forcing
    & 97.02 & 96.09 & 98.33 & 54.97 
    & 60.43 & 68.74  \\

&
FadeMem
    & 97.00 & 96.13 & 98.49 & \textbf{70.29}
    & 61.27 & 71.07 \\

&
\ffcell{\textbf{FrameFold}}
    & \ffcell{\textbf{97.20}}
    & \ffcell{\textbf{96.28}}
    & \ffcell{\textbf{98.62}}
    & \ffcell{68.68}
    & \ffcell{\textbf{61.68}}
    & \ffcell{\textbf{72.18}}
\\

\midrule


\multirow[c]{3}{*}{120s}
&
Deep Forcing
    & 96.85 & 95.94 & 98.25 & 50.96 & 59.71 & 67.95 \\

&
FadeMem
    & 96.85 & 96.03 & 98.43 & 67.96
    & 60.63 & 70.66  \\

& 
\ffcell{\textbf{FrameFold}}
    & \ffcell{\textbf{97.17}}
    & \ffcell{\textbf{96.24}}
    & \ffcell{\textbf{98.62}}
    & \ffcell{\textbf{68.85}}
    & \ffcell{\textbf{61.23}}
    & \ffcell{\textbf{71.92}}
     \\

\midrule


\multirow[c]{3}{*}{240s}
&
Deep Forcing
    & 96.65 & 95.81 & 98.13 & 46.66 
    & 58.78 & 67.35 \\

&
FadeMem
    & 96.76 & 95.97 & 98.40 & 67.42
    & 59.95 & 70.00  \\

&
\ffcell{\textbf{FrameFold}}
    & \ffcell{\textbf{97.12}}
    & \ffcell{\textbf{96.19}}
    & \ffcell{\textbf{98.59}}
    & \ffcell{\textbf{68.73}}
    & \ffcell{\textbf{60.53}}
    & \ffcell{\textbf{71.67}}
     \\

\midrule


\multirow[c]{3}{*}{480s}
&
Deep Forcing
    & 96.33 & 95.72 & 97.68 & 41.93 
    & 58.13 & 65.86  \\

&
FadeMem
    & 96.63 & 95.86 & 98.41 & 67.88
    & 58.61 & 69.68  \\

&
\ffcell{\textbf{FrameFold}}
    & \ffcell{\textbf{97.02}}
    & \ffcell{\textbf{96.12}}
    & \ffcell{\textbf{98.56}}
    & \ffcell{\textbf{69.79}}
    & \ffcell{\textbf{60.01}}
    & \ffcell{\textbf{71.29}}
  \\

\bottomrule
\end{tabular}
\end{center}

\end{table}

\begin{figure}[!htb]
    \centering
    \includegraphics[width=0.94\linewidth]{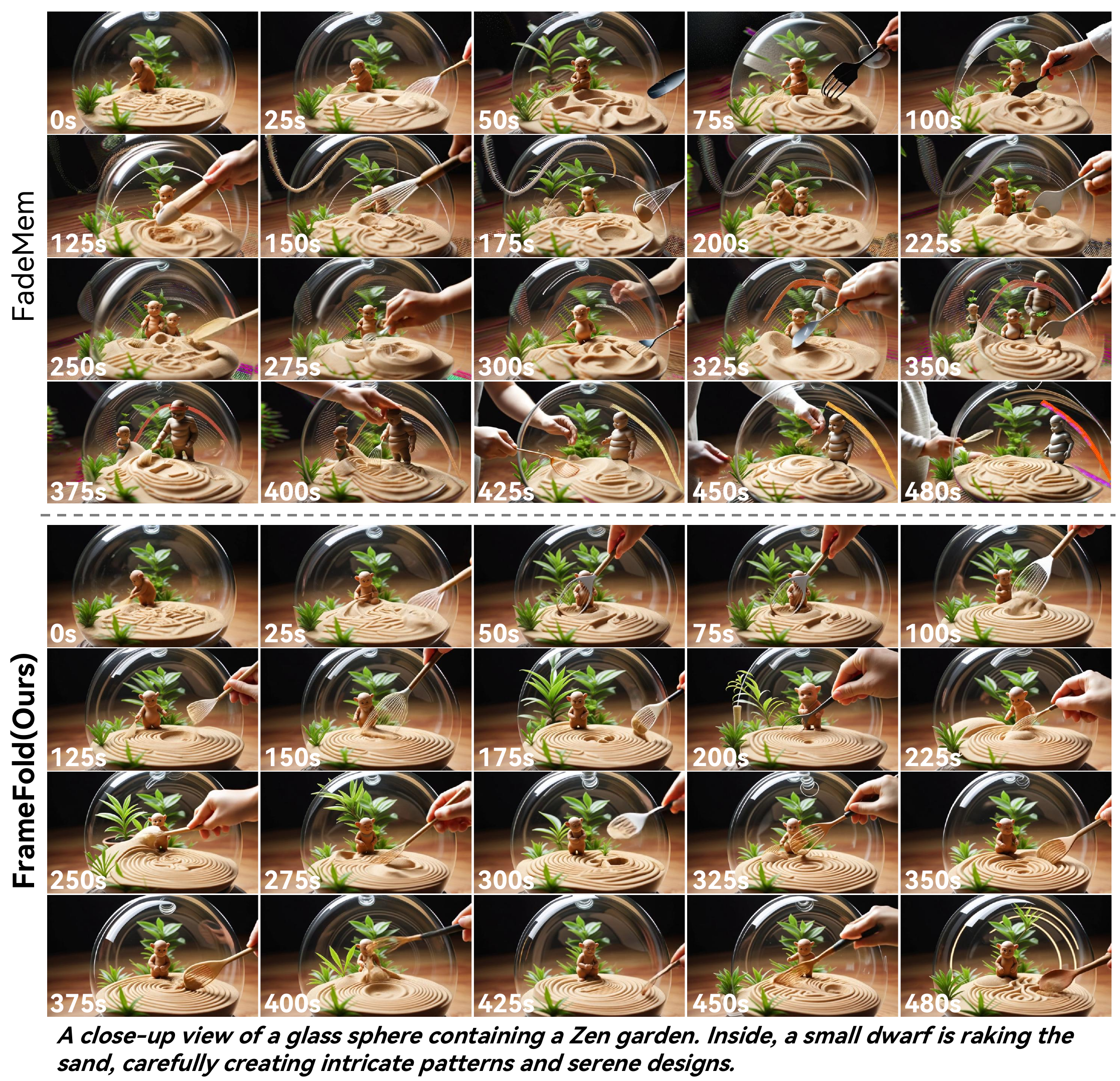}
    \caption{\textbf{Qualitative comparison of \method and FadeMem over
    480 seconds of generation.}
    FadeMem (top) and \method (bottom) use Self Forcing with the same prompt
    and a cache budget of 12 frame-equivalent KV entries.}
    \label{fig:long-duration-full}
\end{figure}

\paragraph{Qualitative results.}
Figure~\ref{fig:long-duration-full} compares subject and scene evolution
under \method and FadeMem over eight minutes of generation, using a garden
scene inside a glass sphere. In the displayed frames, FadeMem introduces
additional figures at 200--250 seconds, followed by further changes in
subject appearance and surrounding structures relative to the initial
scene. \method preserves a single central figure and a more stable
overall layout. The raking tool's position and the sand patterns change
over time, while the central figure remains recognizable in the final
480-second frame. This sequence provides a visual counterpart to the
subject and background consistency results.

\subsection{Supplementary Results: Merge Selection and Value Fitting}
\label{app:merge-visual}

\paragraph{Experimental setup.}
This section supplements the ablation study in Section~\ref{sec:ablations}
with configuration details and qualitative comparisons. We combine three
merge-selection criteria---K diff, V diff, and O diff---with two
value-merging rules, Span and Optim, yielding six configurations. All six
use Self Forcing to generate 30-second videos on 48 prompts, with a cache
budget of 12 frame-equivalent KV entries, a pooling factor of 4, and
queries from the three most recently completed frames for evaluation.
All other generation settings remain fixed.

\paragraph{Ablation variants.}
K diff and V diff rank adjacent pairs by the mean $L_2$ distance between
their pooled key or value features at corresponding spatial locations.
O diff instead measures the conditional-response distortion between the
original pair and its merged representative under the same recent queries.
Each criterion selects the eligible pair with the lowest score. Keys are
merged using temporal-span weights in every variant. Span applies these
same weights to values, whereas Optim holds the merged keys fixed and
fits a shared value-merging coefficient through the response-matching
objective in Equation~\ref{eq:value-fit-objective}. This fitting objective
is identical across all three selection criteria: for example,
K diff + Optim uses key-feature distance for pair selection and
conditional-response matching for value fitting. Pooling is used for
candidate evaluation and coefficient fitting, while the selected merge
is applied to the complete spatial grids. O diff + Optim corresponds to
the full \method configuration.

\begin{figure}[!htb]
    \centering
    \includegraphics[width=0.96\linewidth]{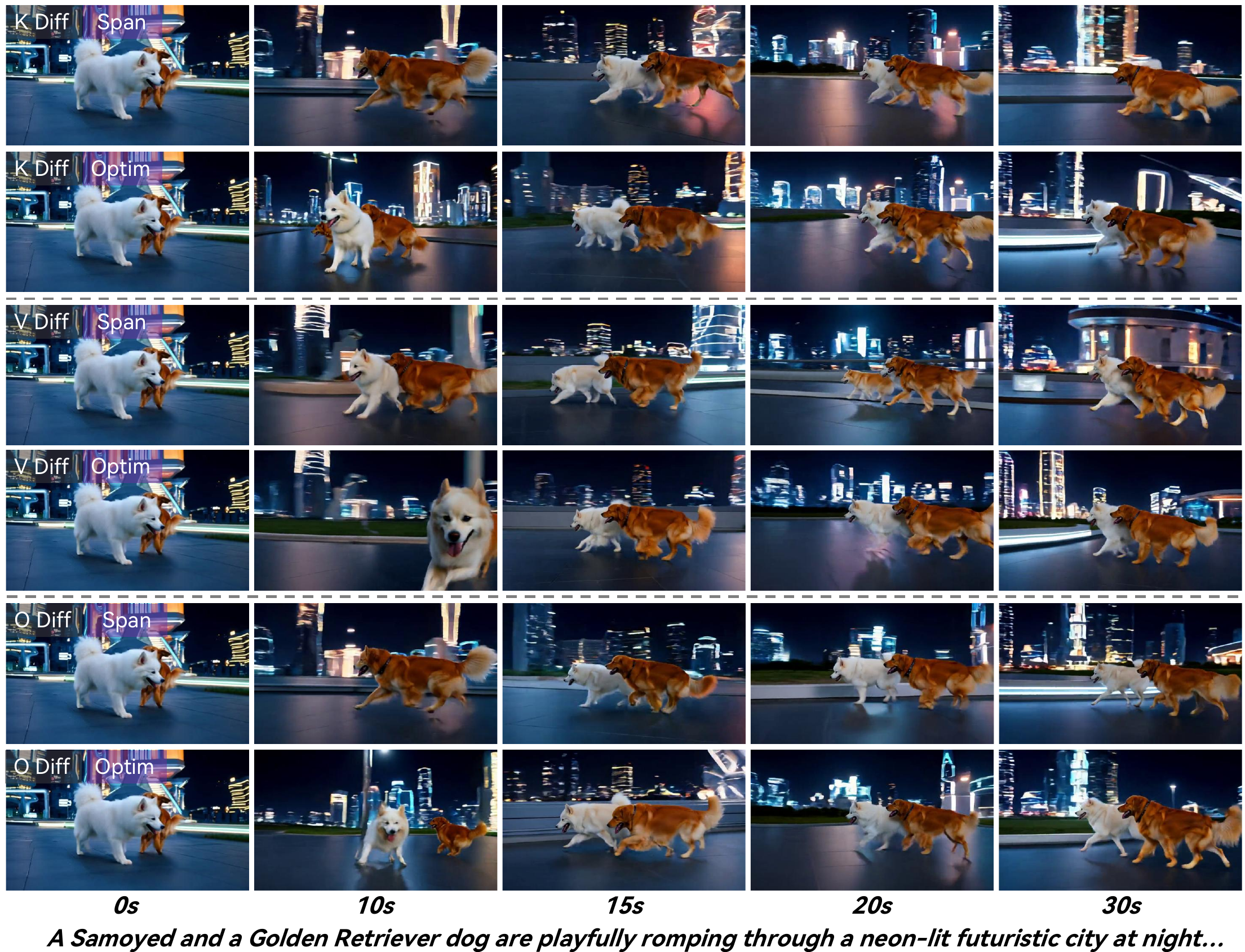}
    \caption{\textbf{Ablation on merge selection and value fitting.}
    Qualitative results are shown for Self Forcing with 30-second videos
    and a cache budget of 12 frame-equivalent KV entries.
    O diff + Optim corresponds to \method.}
    \label{fig:ablation-merge-visual}
\end{figure}

\paragraph{Qualitative results.}
Figure~\ref{fig:ablation-merge-visual} shows the six configurations starting
from the same initial frame. With Optim fixed, the selection criteria
yield different subject depictions: at 10 seconds, V diff primarily shows
a close-up of the white Samoyed, whereas O diff shows both the Samoyed and
the golden retriever. With O diff fixed, the two dogs overlap substantially
in the 10-second frame under Span, whereas Optim shows them separately in
the foreground and background. Under O diff + Optim, the white Samoyed
and golden retriever remain visually distinguishable in subsequent sampled
frames, including the final 30-second frame. These paired comparisons
illustrate differences in subject visibility and composition associated
with merge selection and value construction, providing a visual complement
to the quantitative ablation in Table~\ref{tab:ablation-merge-v}.

\subsection{Supplementary Results: Response Objective}
\label{app:response-visual}

\paragraph{Experimental setup.}
We compare conditional and full-cache response matching for both value
fitting and merge selection, using Self Forcing to generate 30-second
videos on 128 prompts with a cache budget of 12 frame-equivalent KV
entries. Conditional matching compares the block-local responses of the
original pair and its representative. Full-cache matching compares the
attention outputs of the entire cache before and after the candidate
replacement. All other generation and merge settings remain fixed.

\begin{figure}[!htb]
    \centering
    \includegraphics[width=0.95\linewidth]{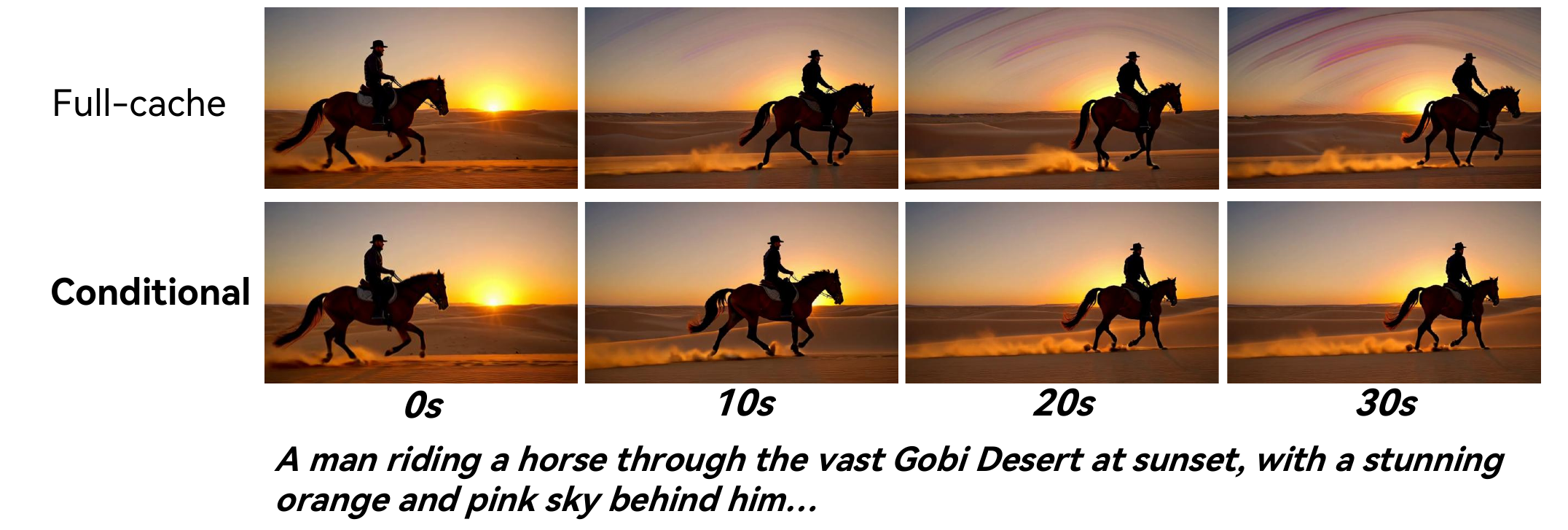}
    \caption{\textbf{Ablation on the response objective.}
    Qualitative results show 30-second Self Forcing generation with a
    cache budget of 12 frame-equivalent KV entries. Full-cache (top) and
    conditional (bottom, \method) use the same prompt; each objective
    is applied to both value fitting and merge selection.}
    \label{fig:ablation-response-visual}
\end{figure}

\paragraph{Qualitative results.}
Figure~\ref{fig:ablation-response-visual} shows a rider crossing a desert
at sunset, complementing the quantitative results in
Table~\ref{tab:ablation-response}. Both variants preserve the rider and
horse across the sampled frames, with differences in the sky's color bands
and the shape of the dust trail at 20 and 30 seconds.

\subsection{Additional Experiments: Evaluation on LongLive}
\label{app:longlive}

\paragraph{Experimental setup.}
To evaluate \method on an additional backbone, we generate 120-second
videos with LongLive~\citep{yang2025longlive} using the same 128 prompts
as the main comparison. We compare the original LongLive model,
Deep Forcing, FadeMem, and \method using the six VBench-Long dimensions.
All methods share a cache budget of 12 frame-equivalent KV entries,
random seeds, and sampling settings. \method uses a pooling factor of 4
and queries from the three most recently completed historical frames;
other settings follow Section~\ref{sec:experimental-setup}.

\paragraph{Quantitative results.}
Table~\ref{tab:main-results-longlive} shows that \method achieves the
highest subject consistency, background consistency, aesthetic quality,
and imaging quality among the compared methods. Background consistency
and aesthetic quality exceed those of the strongest baseline, FadeMem, by 0.36
and 0.56 points, respectively. Motion smoothness is slightly lower than
Deep Forcing's, while dynamic degree is lower than the original LongLive
model's and higher than both Deep Forcing's and FadeMem's. These results extend the
consistency and visual-quality benefits of \method to LongLive at
120 seconds.

\begin{table}[t]
\caption{\textbf{Quantitative comparison on LongLive for 120-second video generation.}
All methods use the same 128 prompts as the main comparison and a cache
budget of 12 frame-equivalent KV entries. We report six VBench-Long
dimensions; higher is better, and the best scores are in \textbf{bold}.}
\label{tab:main-results-longlive}

\setlength{\tabcolsep}{2.5pt}

\begin{center}
\begin{tabular}{
    c
    l
    c c c c c c 
}
\toprule

\textbf{Config}
&
\textbf{Method}
&
\makecell{\textbf{Subject}\\\textbf{Consist.}$\uparrow$}
&
\makecell{\textbf{Background}\\\textbf{Consist.}$\uparrow$}
&
\makecell{\textbf{Motion}\\\textbf{Smooth.}$\uparrow$}
&
\makecell{\textbf{Dynamic}\\\textbf{Degree}$\uparrow$}
&
\makecell{\textbf{Aesthetic}\\\textbf{Quality}$\uparrow$}
&
\makecell{\textbf{Imaging}\\\textbf{Quality}$\uparrow$}

\\

\midrule


\multirow[c]{4}{*}{120s}
&
LongLive
    & 97.24 & 96.22 & 98.43 & \textbf{47.38} & 57.61 & 66.93 \\

&
Deep Forcing
    & 97.58 & 96.25 & \textbf{98.53} & 39.59
    & 57.88 & 67.35  \\

&
FadeMem
    & 97.45 & 96.39 & 98.47 & 45.96 & 58.53 & 67.65 \\

&
\ffcell{\textbf{FrameFold}}
    & \ffcell{\textbf{97.76}}
    & \ffcell{\textbf{96.75}}
    & \ffcell{98.50}
    & \ffcell{46.32}
    & \ffcell{\textbf{59.09}}
    & \ffcell{\textbf{68.10}}
    \\

\bottomrule
\end{tabular}
\end{center}

\end{table}

\begin{figure}[!htb]
    \centering
    \includegraphics[width=0.95\linewidth]{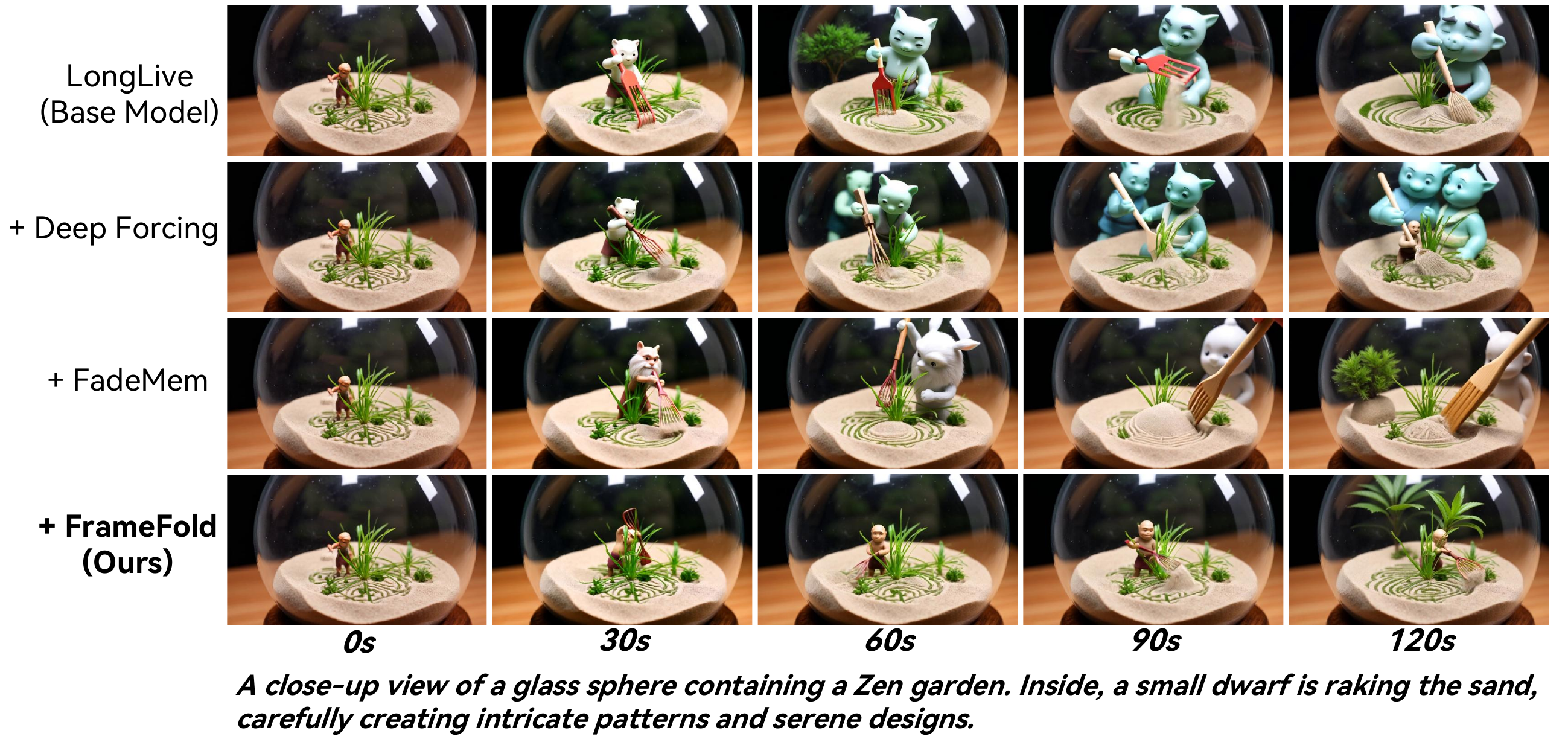}
    \caption{\textbf{Qualitative comparison on LongLive over 120 seconds of generation.}
    From top to bottom: the original LongLive model, Deep Forcing,
    FadeMem, and \method. Frames are shown at 0, 30, 60, 90, and
    120 seconds. All methods use the same prompt and random seed,
    with a cache budget of 12 frame-equivalent KV entries.}
    \label{fig:main-results-longlive}
\end{figure}

\paragraph{Qualitative results.}
Figure~\ref{fig:main-results-longlive} shows a small figure raking a
Zen garden inside a glass sphere. In the original LongLive and FadeMem
sequences, the initial figure changes into larger animal-like characters,
while Deep Forcing introduces multiple figures in later frames.
\method retains a single small humanoid figure across the sampled
timestamps, with changes in the rake position and sand patterns.
This example illustrates the subject-preservation differences reflected
in the quantitative comparison.

\subsection{Additional Experiments: Sensitivity to Cache Budget and Pooling}
\label{app:sensitivity}

\paragraph{Experimental setup.}
This section examines how the cache budget and spatial pooling factor
affect the generation quality of \method. The cache budget controls the
number of stored frame-equivalent KV entries, while the pooling factor
controls the spatial resolution of the proxies used for value fitting and
merge scoring. Both studies use Self Forcing to generate 60-second videos
on 48 prompts. The cache-budget study compares 12, 15, 18, and 21 entries
with the pooling factor fixed at 4. The spatial-pooling study compares
factors of 1, 2, 4, and 8 with the cache budget fixed at 12 entries;
a factor of 1 uses the original spatial resolution.

\paragraph{Cache budget.}
Table~\ref{tab:ablation-cache-budget} shows that \method outperforms
Deep Forcing in subject consistency, background consistency, motion
smoothness, aesthetic quality, and imaging quality at every tested budget,
with imaging-quality gains of 2.05--3.42 points. Using 12 entries,
\method also exceeds Deep Forcing with 21 entries across all six metrics:
it uses 42.9\% fewer cache entries while improving aesthetic quality and
imaging quality by 2.11 and 3.66 points, respectively. For \method,
increasing the budget does not yield monotonic improvements across the
metrics, and different metrics reach their highest scores at different
budgets. Since a consolidated entry can represent multiple historical
frames, the cache budget determines the granularity of the history
representation, and the number of entries does not directly specify the
historical span they cover. These results show that \method maintains its
quality advantage across the tested budgets and support using 12 entries
as a compact default configuration.

\begin{table}[!htb]
\caption{
\textbf{Ablation on KV cache budget.}
Experiments on Self Forcing use 48 prompts and 60-second videos. \method uses
a pooling factor of 4. Cache budgets are measured in frame-equivalent KV
entries; higher scores are better. The default FrameFold configuration is
highlighted in blue.
}
\label{tab:ablation-cache-budget}

\setlength{\tabcolsep}{2.5pt}

\begin{center}
\begin{tabular}{
    l c
    c c c c c c
}
\toprule

\textbf{Method}
&
\makecell{\textbf{Cache}\\\textbf{Budget}}
&
\makecell{\textbf{Subject}\\\textbf{Consist.}$\uparrow$}
&
\makecell{\textbf{Background}\\\textbf{Consist.}$\uparrow$}
&
\makecell{\textbf{Motion}\\\textbf{Smooth.}$\uparrow$}
&
\makecell{\textbf{Dynamic}\\\textbf{Degree}$\uparrow$}
&
\makecell{\textbf{Aesthetic}\\\textbf{Quality}$\uparrow$}
&
\makecell{\textbf{Imaging}\\\textbf{Quality}$\uparrow$}
\\
\midrule

\multirow[c]{4}{*}{Deep Forcing}
& 12
            & 96.75 & 95.98 & 98.09 & 61.32 & 59.63 & 68.85  \\
& 15
            & 96.65 & 95.73 & 98.02 & 66.32 & 58.90 & 68.34 \\
& 18 
            & 96.64 & 95.83 & 97.98 & 61.74 & 58.88 & 68.11 \\
& 21 
            & 96.47 & 95.65 & 97.96 & 59.37 & 58.44 & 67.24 \\

\midrule
\multirow[c]{4}{*}{FrameFold}
& \ffcell{12}
            & \ffcell{97.15} & \ffcell{96.27} & \ffcell{98.27} & \ffcell{67.08} & \ffcell{60.55} & \ffcell{70.90} \\
& 15
            & 97.19 & 96.21 & 98.28 & 64.37 & 60.28 & 70.74 \\
& 18 
            & 96.93 & 96.08 & 98.25 & 68.26 & 60.03 & 70.58 \\
& 21 
            & 96.94 & 96.09 & 98.41 & 64.58 & 59.88 & 70.66 \\

\bottomrule
\end{tabular}
\end{center}

\end{table}

\paragraph{Spatial pooling.}
Spatial pooling constructs the proxies used for value fitting and merge
scoring, and the resulting merge coefficients are applied to the complete
spatial KV grids. Across the four pooling settings, subject consistency,
background consistency, and motion smoothness each vary by at most
0.16 points (Table~\ref{tab:ablation-pooling}). The default factor of 4
achieves the highest background consistency and aesthetic quality, with
imaging quality of 70.90, close to the 70.97 obtained at factor 1. Factor 2
achieves the highest imaging quality of 71.30, while factor 8 reduces it
to 70.37. At factor 4, each historical entry's original $30\times52$ grid
is reduced to an $8\times13$ proxy grid, decreasing the token count from
1,560 to 104, while the stored KV representatives retain their full
spatial resolution. Factor 4 thus maintains similar generation quality
with substantially smaller evaluation proxies, supporting its use as the
default pooling configuration.

\begin{table}[!htb]
\caption{
\textbf{Ablation on spatial pooling factor.}
Experiments on Self Forcing use 48 prompts, 60-second videos, and a cache budget
of 12 frame-equivalent KV entries. A pooling factor of 1 uses the original
spatial resolution. Higher scores are better. The default configuration is
highlighted in blue.
}
\label{tab:ablation-pooling}

\setlength{\tabcolsep}{2.5pt}

\begin{center}
\begin{tabular}{
    l c
    c c c c c c
}
\toprule

\textbf{Backbone}
&
\makecell{\textbf{Pooling}\\\textbf{Factor}}
&
\makecell{\textbf{Subject}\\\textbf{Consist.}$\uparrow$}
&
\makecell{\textbf{Background}\\\textbf{Consist.}$\uparrow$}
&
\makecell{\textbf{Motion}\\\textbf{Smooth.}$\uparrow$}
&
\makecell{\textbf{Dynamic}\\\textbf{Degree}$\uparrow$}
&
\makecell{\textbf{Aesthetic}\\\textbf{Quality}$\uparrow$}
&
\makecell{\textbf{Imaging}\\\textbf{Quality}$\uparrow$}
\\
\midrule

\multirow[c]{4}{*}{Self Forcing}
& 1
            & 97.12 & 96.19 & 98.29 & 66.39 & 60.29 & 70.97  \\
& 2 
            & 97.20 & 96.20 & 98.28 & 65.69 & 60.24 & 71.30 \\
& \ffcell{4}
            & \ffcell{97.15} & \ffcell{96.27} & \ffcell{98.27} & \ffcell{67.08} & \ffcell{60.55} & \ffcell{70.90} \\
& 8 
            & 97.09 & 96.11 & 98.17 & 67.01 & 60.28 & 70.37 \\
                
\bottomrule
\end{tabular}
\end{center}

\end{table}


\subsection{Additional Experiments: Generation Throughput}
\label{app:efficiency}

\paragraph{Experimental setup.}
We measure generation throughput in output frames per second (FPS) using
Self Forcing on NVIDIA H20 hardware, excluding VAE decoding and video
saving from the timing. All four methods use a cache budget of 12
frame-equivalent KV entries, and we additionally evaluate FIFO and
Deep Forcing with 21 entries. \method uses a pooling factor of 4 and
queries from the three most recently completed historical frames for
value fitting and merge scoring. Other settings follow the defaults in
Section~\ref{sec:experimental-setup}.

\begin{figure}[!htb]
    \centering
    \includegraphics[width=0.6\linewidth]{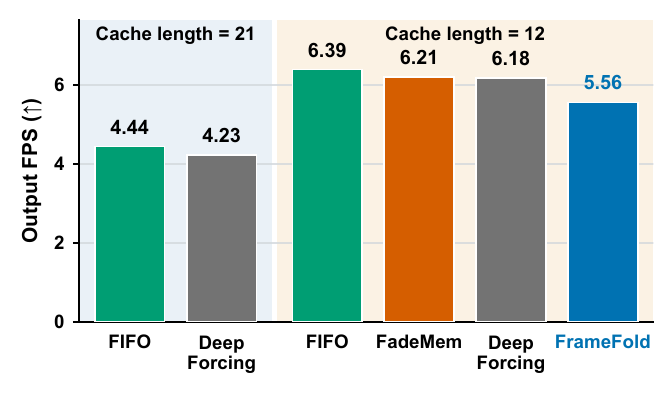}
    \caption{\textbf{Quantitative comparison of generation throughput.}
    Throughput is measured in output frames per second (FPS) on
    Self Forcing with an NVIDIA H20, excluding VAE decoding and video
    saving. All four methods use a cache budget of 12 frame-equivalent KV
    entries; FIFO and Deep Forcing are additionally evaluated with
    21 entries.}
    \label{fig:efficiency}
\end{figure}

\paragraph{Results and analysis.}
At the same budget of 12 entries, \method achieves 5.56 FPS, compared
with 6.39 for FIFO, 6.21 for FadeMem, and 6.18 for Deep Forcing
(Figure~\ref{fig:efficiency}). Its throughput is 13.0\%, 10.5\%, and
10.0\% lower than that of these three methods, respectively. Together with the
quality results at the same budget in Table~\ref{tab:main-results}, these
measurements characterize the quality--throughput trade-off of \method:
it achieves higher consistency and visual quality within a fixed cache
capacity while requiring more generation time. As an additional comparison
across cache configurations, \method with 12 entries achieves
$1.31\times$ the throughput of Deep Forcing with 21 entries and obtains
higher scores across all six dimensions in the cache-budget study
(Table~\ref{tab:ablation-cache-budget}).

\section{Implementation Details}
\label{app:implementation-details}

\subsection{Efficient Implementation of FrameFold}
\label{app:efficient-implementation}

\paragraph{Pooled representations.}
We use adaptive average pooling to construct compact representations for
fitting and scoring. Keys and values are spatially pooled independently
within each historical entry, while queries from the latest $R$ completed
historical frames are jointly pooled over time and space into a single
set of query probes. Our default uses $P=4$ and $R=3$, reducing the
$30\times52$ spatial grid to $8\times13$. This reduces each historical
entry from 1,560 tokens to 104 tokens and produces 104 query probes.
Fitting and scoring use these pooled representations, and the selected
mixing coefficients are applied to the complete spatial grids.

\paragraph{Key merging and position conventions.}
We merge keys before RoPE and add the two entries' temporal spans.
Starting from unit spans, repeated span-weighted merging keeps each
representative's keys equal to the mean of the original frame keys it
covers. For response evaluation, the original keys use their respective
current cache positions, while the merged keys use the left entry's
position. Queries retain the positional encoding from the start of the
cache update. Queries and keys receive RoPE before pooling; values are
pooled directly.

\paragraph{Degenerate value fitting.}
When $D_i=0$ in Section~\ref{sec:structured-memory}, the representative's
response is independent of $\alpha$. We set $\alpha_i^\star=0$, which is a
minimizer of the squared response error in this case.

\paragraph{Batched candidate evaluation.}
At each layer and merge iteration, we organize all adjacent candidate
pairs along a candidate dimension and share the same query probes across
them. Batched tensor operations compute conditional responses, fit
value-mixing coefficients, and score response distortions for all
candidates. Scoring reuses intermediate results from fitting to directly
obtain the fitted representative responses. Each iteration selects and
executes one merge, then evaluates the updated candidate set in the next
batch of computations. Query probes remain fixed throughout the cache
update.

Algorithm~\ref{alg:framefold-cache-update} summarizes one cache update,
performed independently at each layer. Each entry stores
$(K^{\mathrm{pre}},V,s)$, and the budget includes the current chunk.
The budget must accommodate all protected entries and at least one
representative of the candidate history.

\begin{algorithm}[!htb]
\caption{\method cache consolidation for one layer}
\label{alg:framefold-cache-update}
\begin{algorithmic}[1]
\Require Ordered cache $\mathcal H$ including the current chunk, total budget
$B$, recent historical queries $Q$ encoded at the start of the update,
pooling factor $P$
\Ensure Updated cache $\mathcal H$ with $|\mathcal H|\leq B$
\State Fix the protected sink, recent-history, and current-chunk entries
\State $\widehat Q\gets\operatorname{PoolQueries}(Q,P)$
\While{$|\mathcal H|>B$}
    \State $\mathcal E\gets$ left-entry indices of adjacent pairs in the unprotected region
    \State Rebuild pooled KV proxies at the entries' current cache positions $p_j$
    \ForAll{$i\in\mathcal E$ \textbf{in parallel}}
        \State $\beta_i\gets s_i/(s_i+s_{i+1})$
        \State $\widehat{\widetilde K}_i\gets
        \operatorname{Pool}_P\!\bigl(\operatorname{RoPE}_{p_i}
        (\beta_iK_i^{\mathrm{pre}}+(1-\beta_i)K_{i+1}^{\mathrm{pre}})\bigr)$
        \State Compute $\widehat O_i^-$ and $A_i^+$ under $\widehat Q$ using block-local attention
        \State $C_i\gets A_i^+\widehat V_{i+1}$;
        $D_i\gets A_i^+(\widehat V_i-\widehat V_{i+1})$
        \State Fit $\alpha_i^\star$ using Equation~\ref{eq:value-fit-solution};
        set $\alpha_i^\star=0$ if $D_i=0$
        \State $\widehat O_i^\star\gets C_i+\alpha_i^\star D_i$
        \State $c_i\gets(n_hU)^{-1}\sum_{h,u}
        \|\widehat O^-_{i,hu}-\widehat O^\star_{i,hu}\|_2$
    \EndFor
    \State $i^\star\gets\arg\min_{i\in\mathcal E}c_i$
    \State $\widetilde K^{\mathrm{pre}}\gets
    \beta_{i^\star}K_{i^\star}^{\mathrm{pre}}
    +(1-\beta_{i^\star})K_{i^\star+1}^{\mathrm{pre}}$
    \State $\widetilde V\gets\alpha_{i^\star}^\star V_{i^\star}
    +(1-\alpha_{i^\star}^\star)V_{i^\star+1}$
    \State Replace entries $i^\star,i^\star+1$ by
    $(\widetilde K^{\mathrm{pre}},\widetilde V,s_{i^\star}+s_{i^\star+1})$
    \State Update cache positions
\EndWhile
\State \Return $\mathcal H$
\end{algorithmic}
\end{algorithm}

\subsection{Baseline Configurations}
\label{app:baseline-configurations}

\paragraph{Shared cache conventions.}
All baselines follow the cache-budget convention in
Section~\ref{sec:experimental-setup}: the default budget is 12
frame-equivalent KV entries, including the three-entry current chunk,
with three initial sink frames and the three most recent historical
frames protected. Other budgets are specified in the corresponding
experiments. Within each backbone, all methods use the same random seed
and sampling settings for each prompt, ensuring identical generated
content before the first cache eviction or merge and isolating the
effect of the KV cache management policy on generation results.

\paragraph{Original backbones.}
Self Forcing and Causal Forcing use their original FIFO cache eviction
policies and are listed under their model names in
Table~\ref{tab:main-results}.

\paragraph{Deep Forcing.}
We use the official Deep Forcing implementation~\citep{yi2025deepforcing}.
Each eviction event removes a number of tokens equivalent to three
complete spatial grids.

\paragraph{FadeMem.}
We implement FadeMem's inference-time memory policy~\citep{lu2026fademem}
in the MemRoPE framework~\citep{kim2026memrope}, following the temporal
allocation and merge rules in
\href{https://arxiv.org/abs/2606.10671v1}{arXiv version~1 (v1)} of FadeMem.
We set the temporal allocation exponent to $\beta=0.3$ and merge both
keys and values using temporal-span weights.

\section{Discussion}
\label{app:discussion}

\subsection{Limitations}
\label{app:limitations}

\paragraph{Future query demands.}
\method fits and evaluates representatives using recent historical
queries, without explicitly modeling changes in future query demands.
Query-proxy construction and conditional-response matching can be
designed separately. Future Forcing~\citep{luo2026futureforcing} constructs
future query proxies from historical statistics, offering a direction for
extending our evaluation probes.

\paragraph{Additional computational cost.}
Fitting and scoring candidate representatives introduce additional
computation. Pooling, closed-form fitting, and batched evaluation reduce
this cost, but the current implementation has lower throughput than the
compared methods under the same cache budget
(Appendix~\ref{app:efficiency}). Latency-sensitive applications therefore
require balancing gains in consistency and visual quality against
generation speed. Improving the execution efficiency of candidate
evaluation is a direction for future optimization.

\end{document}